\documentclass[conference]{IEEEtran}
\IEEEoverridecommandlockouts
\usepackage{times}
\usepackage{subfigure}
\usepackage[table,xcdraw,dvipsnames]{xcolor}
\usepackage{multirow}
\usepackage{mathtools}
\usepackage{outlines}
\usepackage{times}
\usepackage{breqn}
\usepackage{amsmath,amsthm,amssymb,accents,mathtools,amsfonts} 
\usepackage{algorithm}
\usepackage{algorithmic}
\usepackage{makecell}
\usepackage[font=footnotesize]{caption}
\usepackage{subcaption}
\usepackage[bb=boondox]{mathalfa}
\usepackage{graphicx}
\usepackage{epsfig}
\usepackage{authblk}
\usepackage{footnote}
\usepackage{xcolor}
\usepackage{booktabs}
\usepackage{ragged2e}  

\usepackage[numbers]{natbib}

\usepackage{multicol}
\usepackage[colorlinks=true,linkcolor=black,anchorcolor=black,citecolor=black,filecolor=black,menucolor=black,runcolor=black,urlcolor=black]{hyperref}

\DeclareMathOperator*{\argmin}{arg\,min}

\begin{document}

\newif\ifcommentson
\commentsontrue
\newcounter{BohaoCount}
\addtocounter{BohaoCount}{1}
\newcommand{\bohao}[1]{\textcolor{Green}{\ifcommentson\textbf{(\theBohaoCount) BZ}: \textbf{(#1)}\fi}\addtocounter{BohaoCount}{1}}

\newcounter{DanielCount}
\addtocounter{DanielCount}{1}
\newcommand{\Dan}[1]{\textcolor{Orange}{\ifcommentson\textbf{(\theDanielCount) DH}: \textbf{(#1)}\fi}\addtocounter{DanielCount}{1}}

\newcounter{RamCount}
\addtocounter{RamCount}{1}
\newcommand{\Ram}[1]{\textcolor{Red}{\ifcommentson\textbf{(\theRamCount) RV-M}: \textbf{(#1)}\fi}\addtocounter{RamCount}{1}}

\newcounter{FixCount}
\addtocounter{FixCount}{1}
\newcommand{\fix}[1]{\textcolor{Purple}{\ifcommentson\textbf{(\theFixCount) FIX}: (#1)\fi}\addtocounter{FixCount}{1}}

\newtheorem{defn}{Definition}
\newtheorem{rem}[defn]{Remark}
\newtheorem{lem}[defn]{Lemma}
\newtheorem{prop}[defn]{Proposition}
\newtheorem{assum}[defn]{Assumption}
\newtheorem{ex}[defn]{Example}
\newtheorem{thm}[defn]{Theorem}
\newtheorem{cor}[defn]{Corollary}
\newtheorem{con}[defn]{Conjecture}
\newtheorem{problem}[defn]{Problem}

\providecommand{\R}{\ensuremath \mathbb{R}}
\providecommand{\IR}{\ensuremath \mathbb{IR}}
\providecommand{\N}{\ensuremath \mathbb{N}}
\providecommand{\Q}{\ensuremath \mathcal{Q}}
\providecommand{\A}{\ensuremath \mathcal{A}}
\providecommand{\U}{\ensuremath \mathcal{U}}
\providecommand{\K}{\ensuremath \mathcal{K}}
\newcommand{\unitcircle}{\mathbb{S}^1}

\newcommand{\ubar}[1]{\underaccent{\bar}{#1}}

\newcommand{\regtext}[1]{\mathrm{\textnormal{#1}}}
\newcommand{\ol}[1]{\overline{#1}}
\newcommand{\ul}[1]{\underline{#1}}
\newcommand{\defemph}[1]{\emph{#1}}
\newcommand{\ts}[1]{\textsuperscript{#1}}

\newcommand{\comp}{^{\regtext{C}}}
\newcommand{\card}[1]{\left\vert#1\right\vert}
\newcommand{\proj}{\regtext{proj}}
\newcommand{\norm}[1]{\left\Vert#1\right\Vert}
\newcommand{\abs}[1]{\left\vert#1\right\vert}
\newcommand{\pow}[1]{\mathcal{P}\!\left(#1\right)}
\newcommand{\diag}[1]{\regtext{diag}\!\left(#1\right)}
\newcommand{\eig}[1]{\regtext{eig}\!\left(#1\right)}
\newcommand{\union}{\bigcup}
\newcommand{\intersection}{\bigcap}
\newcommand{\trans}{^\top}
\newcommand{\inv}{^{-1}}
\newcommand{\pinv}{^{\dagger}}
\newcommand{\sign}{\regtext{sign}}
\newcommand{\expm}{\regtext{exp}}
\newcommand{\logm}{\regtext{log}}
\newcommand{\skw}{_{\times}}
\newcommand{\bigO}{\mathcal{O}}
\newcommand{\bdry}[1]{\regtext{bd}\!\left(#1\right)}
\renewcommand{\ker}[1]{\regtext{ker}\!\left(#1\right)}
\newcommand{\convhull}[1]{\regtext{CH}\!\left(#1\right)}

\newcommand{\lbl}[1]{_{\regtext{#1}}}
\newcommand{\lo}{\lbl{lo}}
\newcommand{\hi}{\lbl{hi}}

\newcommand{\emptyarr}{[\ ]}
\newcommand{\zeros}{\textit{0}}
\newcommand{\ones}{\textit{1}}
\newcommand{\eye}{\regtext{\textit{I}}}

\newcommand{\interval}[1]{[ #1 ]}
\newcommand{\iv}[1]{[ #1 ]}
\newcommand{\nom}[1]{#1}
\newcommand{\setop}[1]{{\mathrm{\texttt{#1}}}}
\newcommand{\lb}[1]{\underline{#1}}
\newcommand{\ub}[1]{\overline{#1}}

\newcommand{\qA}{q_A(t)}
\newcommand{\Wp}{W_{\text{passive}}}

\providecommand{\R}{\ensuremath \mathbb{R}}
\newcommand{\plan}{_p}
\newcommand{\prev}{\lbl{prev}}

\newcommand{\zi}{z_i}
\newcommand{\zj}{z_j}
\newcommand{\rbf}{\mathbf{r}(t)}

\newcommand{\bH}{H}
\newcommand{\Hq}{H(\q, \theta)}
\newcommand{\Hqdot}{\dot{H}(\q, \theta)}
\newcommand{\Haqt}{H_a(\q, \theta)}

\newcommand{\bC}{C}
\newcommand{\Cq}{C(\q, \qd, \theta)}

\newcommand{\bG}{g}
\newcommand{\Gq}{g(\q, \theta)}

\newcommand{\intparams}{[\theta]}
\newcommand{\nomparams}{\theta_0}
\newcommand{\trueparams}{\theta}

\newcommand{\intparamsend}{[\theta_{\text{\footnotesize{e}}}]}
\newcommand{\trueparamsend}{\theta_{\text{\footnotesize{e}}}}
\newcommand{\nomparamsend}{\theta_{\text{\footnotesize{e}}, 0}}
\newcommand{\intparamsrobot}{[\theta_{\text{\footnotesize{r}}}]}
\newcommand{\trueparamsrobot}{\theta_{\text{\footnotesize{r}}}}
\newcommand{\nomparamsrobot}{\theta_{\text{\footnotesize{r}}, 0}}
\newcommand{\unom}{u_{\text{\footnotesize{nom}}}(\qA, \nomparams)}
\newcommand{\urob}{u_{\text{\footnotesize{fb}}}(\qA, \nomparams, \intparams)}
\newcommand{\wdist}{w(\qA, \nomparams, \trueparams)}

\newcommand{\Gqt}{G(\q)}
\newcommand{\GTqt}{G^T(\q)}

\newcommand{\Kr}{K_r}

\makeatletter
\newcommand{\smalloplus}{\mathbin{\mathpalette\make@small\oplus}}
\newcommand{\smallotimes}{\mathbin{\mathpalette\make@small\otimes}}

\newcommand{\lambdamin}{\lambda_h}
\newcommand{\lambdamax}{\lambda_H}
\newcommand{\sigmin}{\sigma_{h}}
\newcommand{\sigmax}{\sigma_{H}}
\newcommand{\ultbound}{\sqrt{\frac{2 V_M}{\sigmin}}}

\newcommand{\qgoal}{q\lbl{goal}}
\newcommand{\qstart}{q\lbl{start}}

\newcommand{\normrho}{||\rho([\Phi])||}
\newcommand{\normwmax}{||w_M||}
\newcommand{\wmax}{w_M}
\newcommand{\wmaxj}{w_{M, j}}

\newcommand{\q}{q(t)}
\newcommand{\dq}{\dot{q}(t)}
\newcommand{\ddq}{\ddot{q}(t)}

\newcommand{\qd}{q_d(t)}
\newcommand{\dqd}{\dot{q}_d(t)}
\newcommand{\ddqd}{\ddot{q}_d(t)}

\newcommand{\dqm}{\dot{q}_{m}(t)}
\newcommand{\ddqm}{\ddot{q}_{m}(t)}

\providecommand{\inertialparams}{\theta_{\text{\footnotesize{ip}}}}
\providecommand{\frictionparams}{\theta_{\text{\footnotesize{f}}}}
\providecommand{\inertialparamsj}{\theta_{\text{\footnotesize{ip}},j}}
\providecommand{\frictionparamsj}{\theta_{\text{\footnotesize{f}},j}}
\providecommand{\inertialparamsend}{\theta_{\text{\footnotesize{e}}}}
\providecommand{\inertialparamsrobot}{\theta_{\text{\footnotesize{r}}}}

\newcommand{\qlim}{q_{j,\regtext{lim}}}
\newcommand{\dqlim}{\dot{q}_{j,\regtext{lim}}}
\newcommand{\ddqlim}{\ddot{q}_{j,\regtext{lim}}}
\newcommand{\taulim}{\tau_{j,\regtext{lim}}}

\newcommand{\nObs}{n_{O}}
\newcommand{\obsset}{{O}}

\newcommand{\FO}{\regtext{\small{FO}}}

\newcommand{\meas}{\mathbf{m}}


\title{Provably-Safe, Online System Identification}


\author{Bohao Zhang$^1$, Zichang Zhou$^1$ and Ram Vasudevan$^1$
\thanks{$^{1}$Robotics Institute, University of Michigan, Ann Arbor, MI \texttt{<jimzhang, zichang, ramv>@umich.edu}.}
\thanks{This work was supported by AFOSR MURI
FA9550-23-1-0400.}}

\maketitle

\begin{abstract}
Precise manipulation tasks require accurate knowledge of payload inertial parameters.
Unfortunately, identifying these parameters for unknown payloads while ensuring that the robotic system satisfies its input and state constraints while avoiding collisions with the environment remains a significant challenge.
This paper presents an integrated framework that enables robotic manipulators to safely and automatically identify payload parameters while maintaining operational safety guarantees. 
The framework consists of two synergistic components: 
an online trajectory planning and control framework that generates provably-safe exciting trajectories for system identification that can be tracked while respecting robot constraints and avoiding obstacles and a robust system identification method that computes rigorous overapproximative bounds on end-effector inertial parameters assuming bounded sensor noise. 
Experimental validation on a robotic manipulator performing challenging tasks with various unknown payloads demonstrates the framework's effectiveness in establishing accurate parameter bounds while maintaining safety throughout the identification process.
The code is available at our project webpage: \href{https://roahmlab.github.io/OnlineSafeSysID/}{https://roahmlab.github.io/OnlineSafeSysID/}.
\end{abstract}

\IEEEpeerreviewmaketitle

\section{Introduction}
\label{sec-01-intro}

Robotic arms have widespread applications in cooperative human-robot environments, including manufacturing, package delivery services, and in-home care. 
These scenarios frequently involve manipulating payloads with uncertain properties while operating under physical constraints. 
To ensure safe operation, robots must avoid collisions while respecting joint position, velocity, and torque limits.
There are three challenges to the safe deployment of robotic arms handling unknown payloads.

\begin{figure}[t!]
    \centering
    \includegraphics[width=\columnwidth]{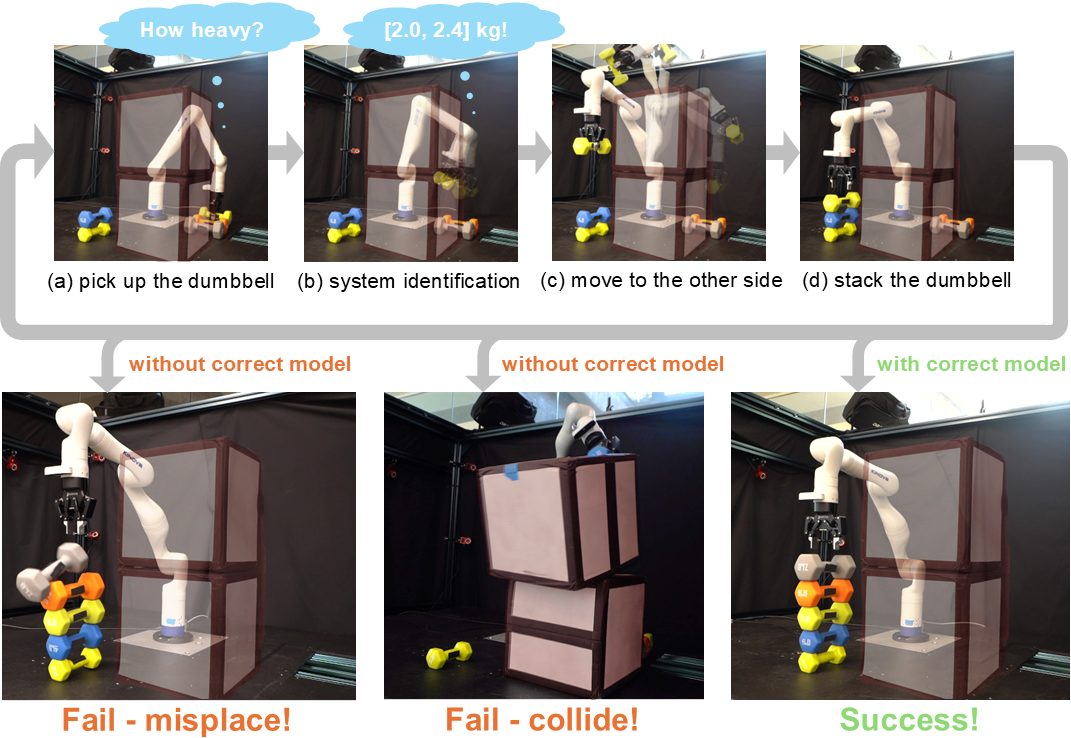}
    \caption{
   This figure illustrates how the method proposed in this paper. 
    (a) Initially, the 7 degree-of-freedom Kinova-gen3 robotic arm picks up a series of heavy dumbbells that are close to the design limit of the robot.
    The inertial parameters of this payload are unknown. 
    (b) The robot then performs online system identification to estimate an interval bound using the method developed in this paper.
    The interval estimate of the inertial parameters generated by the algorithm is guaranteed to include the true inertial parameters of the dumbbell.
    Notably, the data used to compute this interval estimate is constructed in a manner that is guaranteed to be collision free while satisfying joint and torque limits. 
    The inertial parameter bound is then used to update the planner and the controller, which allows the robot to (c) safely move all dumbbells to the other side around the obstacles and then (d) stack them vertically in order of increasing weight, which requires high precision.
    Our experiments illustrate that state-of-the-art methods that do not incorporate such provably overapproximative estimates of the inertial parameters result in a failure to complete the task safely, due to exceeding the torque limits, colliding with obstacles, or misplacing the dumbbells.}
    \label{fig:hardware_demo}
\end{figure}

The first challenge lies in bridging the gap between safe planning and safe control. 
Current research in motion planning and manipulation often overlooks controller tracking performance, where model uncertainties can cause deviations between desired and actual trajectories. 
These deviations may lead to obstacle collisions or torque limit violations.
The second challenge involves accurate estimation of model uncertainties.
For unknown payloads, online system identification becomes necessary to estimate the range of inertial parameters. 
However, larger uncertainty ranges necessitate aggressive controller behavior to guarantee tracking performance, resulting in conservative motion planning which can limit performance.
Therefore, obtaining tight bounds on model uncertainties is crucial for ensuring satisfactory task performance.
The third challenge concerns safety assurances during system identification. 
While extensive research exists on exciting trajectory design for data collection and identification accuracy, these studies primarily focus on industrial robots without payloads and do not consider operational safety during data collection. 
Current approaches rarely address obstacle avoidance or torque limit compliance during the identification process.
This is particularly important when considering online operation wherein the model parameters of a payload may not be known before it is manipulated.

\begin{figure*}[t!]
    \centering
    \includegraphics[width=2.0\columnwidth]{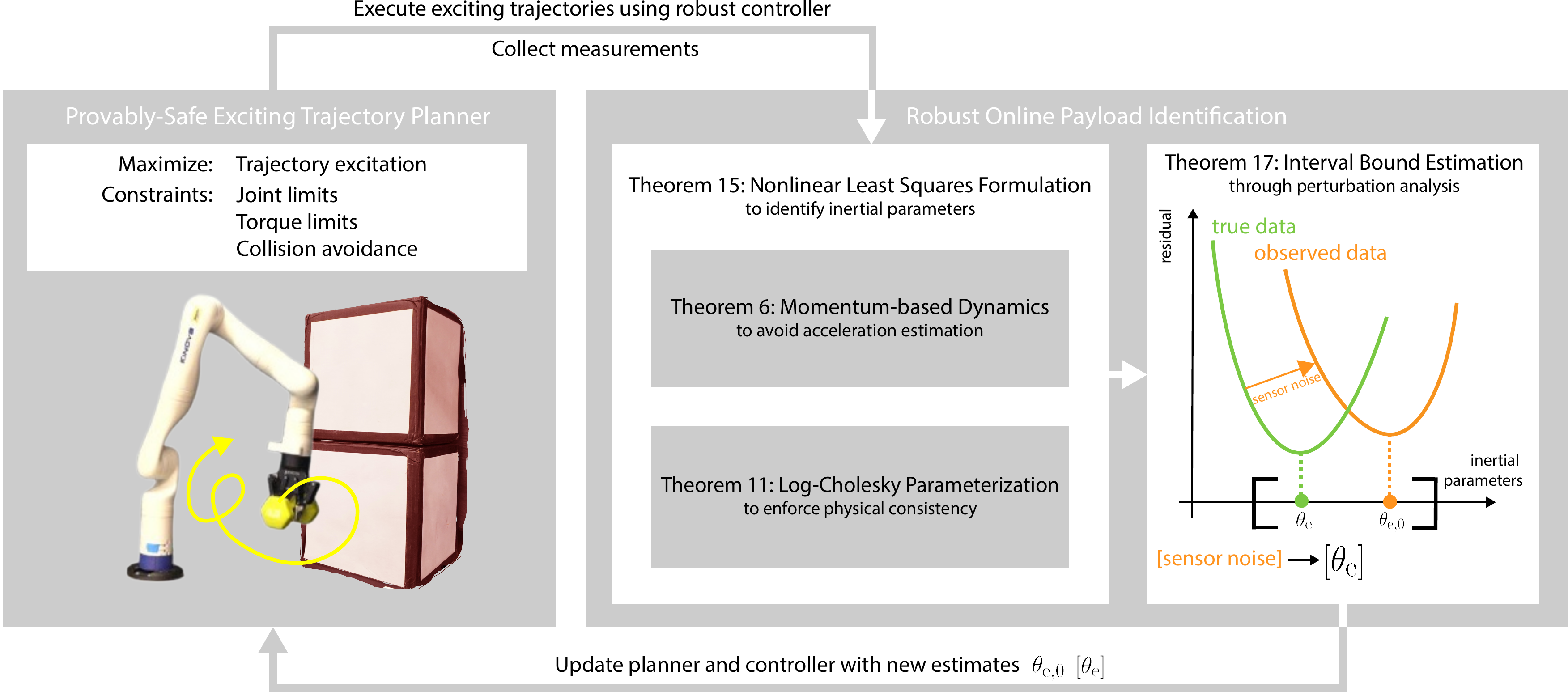}
    \caption{
    This figure summarizes the proposed framework. 
    Initially, the approach assumes an overapproximated bound on the inertial parameters of the robot end-effector with unknown payloads. 
    A trajectory planner (Section \ref{sec-06-trajopt}) then generates a provably-safe, locally exciting desired trajectory in real-time based on this initial bound. 
    A robust controller, modified from \cite[Section VII]{michaux2023cant}, tracks this exciting trajectory while collecting robot data including joint positions, velocities, and applied torques. 
    Using this collected data, the robust system identification method (Section \ref{sec-05-sysid}) generates a new, tighter overapproximated bound on the end-effector's inertial parameters. 
    This process iterates continuously, with additional data enabling more precise parameter estimation and improved planner and controller performance.}
    \label{fig:framwork_overview}
\end{figure*}

This paper presents a provably-safe and real-time system identification framework that addresses these three challenges. 
As illustrated in Figure \ref{fig:framwork_overview}, the proposed framework comprises two integrated components. 
Initially, the approach assumes an overapproximated bound on the inertial parameters of the robot end-effector with unknown payloads. 
A trajectory planning and control framework then generates a provably-safe, locally exciting desired trajectory in real-time based on this initial bound.
This exciting trajectory can be tracked in a provably safe manner.
Using this collected data, the robust system identification method generates a new, tighter, provably overapproximative bound on the end-effector's inertial parameters. 
This process iterates continuously, with additional data enabling more precise parameter estimation and improved planner and controller performance.                       

The key contributions of this paper are two-fold:
First, a real-time, provably-safe trajectory optimization framework that generates locally exciting trajectories for system identification while respecting robot limits and avoiding obstacles.
Second, a robust system identification method that provides an overapproximated bound of the end-effector's inertial parameters given an overapproximated bound of sensor noise.

The remainder of this paper is organized as follows: 
Section \ref{sec-02-related_work} reviews related work. 
Section \ref{sec-03-prelim} introduces relevant notation, mathematical objects, and robot kinematics and dynamics. 
The paper then presents the framework's components in reverse order: 
Section \ref{sec-05-sysid} proposes the system identification method based on momentum regressor, sensitivity analysis, and interval arithmetic. 
Section \ref{sec-06-trajopt} details the trajectory optimization problem formulation. 
Finally, Section \ref{sec-07-exp} demonstrates the method's efficacy through real-world experiments and comparisons with state-of-the-art approaches. 
\section{Related Work}
\label{sec-02-related_work}


The identification of uncertain payloads attached to the end-effector of robotic manipulators leverages the fundamental property that inertial parameters exhibit linearity in the equations of motion \cite{paper-regressor}, enabling the use of least-squares estimation methods \cite{An1985} and, more robustly, weighted least-squares techniques \cite{paper-payload-1, paper-payload-wls}.
Estimated payload parameters are widely applied in various domains, including precise control \cite{paper-adaptive-controller-precise} and collision detection \cite{paper-payload-2}. 
Recent advancements have explored alternative formulations of system dynamics based on momentum \cite{paper-momentum} or energy \cite{paper-energy-regressor}, avoiding estimation of acceleration.
Beyond joint encoders, proprioceptive sensors mounted on the end-effector have been employed to estimate payload properties using filtering-based methods \cite{paper-tactile}. 
Meanwhile, deep learning approaches have gained traction, leveraging neural networks to learn system dynamics directly from system state data \cite{paper-lagrangian1, paper-hamiltonian} or from pure vision input \cite{paper-payload-vision}.

Another important direction relevant to both identification and control of unknown systems is adaptive control. 
Early adaptive controllers aimed to compensate for nonlinear dynamics while decoupling joint interactions \cite{tomizuka1984model, paper-adaptive-review}. 
Subsequent theoretical developments leveraged the linear parameterization of robot dynamics and Lyapunov-based stability analysis to ensure convergence guarantees \cite{paper-adaptive-controller}. 
The introduction of parameter error dynamics through dynamic filtering and torque prediction \cite{slotine1989composite} enabled online parameter adaptation with global stability properties. 
More recent work further advanced this approach by leveraging momentum-based regressors \cite{paper-adaptive-momentum}, thereby eliminating the need for acceleration estimation.
Despite these advancements, adaptive control still faces important limitations. Performance remains highly dependent on persistent excitation, which affects both the convergence speed and steady-state accuracy. 
More critically, classical adaptive control methods typically do not offer explicit guarantees on convergence rate for tracking errors. 
Additionally, they are unable to enforce state and input constraints. 
Although recent research \cite{paper-adaptive-limits} has attempted to address these issues, the proposed solutions are limited to systems with just a single linear actuators and does not generalize to robots governed by nonlinear rigid body dynamics.

For all of these aforementioned methods, the design of proper experiments for data collection, known as exciting trajectory design, has emerged as a critical aspect of system identification \cite{Gautierb,Venture2009,Ayusawa2017,Pressea,Bonnet2016}.
Early methods focused on minimizing the condition number of the regressor matrix \cite{Gautierb}, though later work demonstrated greater effectiveness in optimizing the condition number of sub-regressors \cite{Venture2009}.
Most approaches employ trigonometric series (Fourier series) \cite{paper-iterative-sysid} or B-splines \cite{paper-exciting-1} as trajectory bases to facilitate the enforcement of state and velocity limits, while some methods explore deep learning-based techniques to maximize the Fisher information of the payload properties \cite{paper-asid}.
However, safety guarantees during the identification process are largely unexplored and can be categorized into two main challenges:
(1) Generating a safe exciting trajectory and (2) safely tracking a given trajectory.
To address the first challenge, state of the art approaches enforce only joint and velocity limits.
This may still produce trajectories that violate torque limits.
To the author's best knowledge only \cite{farsoni2019safety} addresses the challenge of generating an exciting trajectory while satisfying safety constraints. 
However, \cite{farsoni2019safety} employs sampling-based planners to generate collision-free trajectories which are unable to guarantee safety between sampled points.
To address the second challenge, state of the art approaches rely on PID controllers \cite{paper-iterative-sysid} or manufacturer-embedded controllers \cite{paper-payload-2}.
As a result, trajectory or controller fine-tuning is often required to balance the trade-off between respecting torque limits and achieving low tracking errors.

To overcome these challenges, the concept of closed-loop state/input sensitivity metric was introduced in \cite{paper-robust-trajectory-planning} to evaluate the impact of model uncertainties on robot behavior.
By designing an optimized feedforward reference trajectory that minimizes these sensitivity metrics, robustness against model uncertainties is inherently embedded in the reference trajectory itself, eliminating the need for specific control strategies \cite{paper-sensitivity}.
This approach has been applied to UAVs \cite{paper-robust-trajectory-planning} and quadrotors \cite{paper-robust-trajectory-planning-1, paper-robust-trajectory-planning-2}, providing a trajectory planning and control framework that considers state/torque limits and obstacle avoidance.
Recent developments have extended this method to robotic manipulators \cite{paper-sensitivity}.
However, this work only considers a narrow range of payload parameters \cite[Section IV]{paper-sensitivity} and has not yet been integrated with collision avoidance.

\section{Preliminaries}
\label{sec-03-prelim}

This section establishes the mathematical foundations necessary for developing the safe system identification framework. 

\subsection{Interval Arithmetic}
\label{sec-03-prelim-02-interval}

To handle uncertainty in robot inertial parameters, this paper employs multidimensional intervals and interval arithmetic. 
For vectors in $\R^n$, we first establish notation: $x_i$ denotes the $i\ts{th}$ component of vector $x$, and for vectors $x,y \in \R^n$, we write $x \preceq y$ when $x_i \leq y_i$ holds for all components $i$.
\begin{defn}[Multidimensional Interval] \label{defn:interval}
For vectors $\ubar{x}, \bar{x} \in \R^n$, a multidimensional interval is the set:
\begin{equation}
[\ubar{x},\bar{x}]:=\{x\in\R^n\ |\ \ubar{x}\preceq x \preceq\bar{x}\},
\end{equation}
where $\ubar{x}$ and $\bar{x}$ serve as the interval's infimum and supremum, respectively.
\end{defn}

The set of all $n$-dimensional intervals is denoted as $\mathbb{I}\mathbb{R}^n$.
For analyzing functions over intervals, we introduce interval evaluation:
\begin{defn}[Interval Evaluation]
For a function $f:\R^n\rightarrow\R^m$ and interval $[\ubar{x},\bar{x}]$, the interval evaluation is defined as:
\begin{equation}
\label{eq-interval-function}
f([\ubar{x},\bar{x}]):= \{f(x)\ |\ x\in[\ubar{x},\bar{x}]\}.
\end{equation}
\end{defn}

\subsection{Robotic Manipulator Model and the Environment}
\label{sec-03-prelim-03-dynamics}



\subsubsection{Dynamics Parameters}
For a fully-actuated robotic system with $n$ joints, let $q:[0,\infty)\in\Q$ denote the generalized trajectory, where $\Q \subset \R^{n}$ represents the robot configuration space. 
For convenience, we let $N_q = \{1,\ldots,n\}$.
The system's behavior is characterized by its \defemph{dynamics parameters}, $\theta \in \R^{14n}$, which comprise two distinct components: inertial parameters and friction parameters.
The inertial parameters $\inertialparams \in \R^{10n}$ describe the mass properties of each link $j$:
\begin{multline}
\inertialparamsj=[m_j, p_{x,j} \cdot m_j, p_{y,j} \cdot m_j,p_{z,j} \cdot m_j, \\
XX_j, YY_j, ZZ_j, XY_j, YZ_j, XZ_j]^T,
\end{multline}
where $m_j$ represents mass, $[p_{x,j} \cdot m_j, p_{y,j} \cdot m_j,p_{z,j} \cdot m_j]$ encodes the first moments of mass, and the remaining six components define the inertia tensor. 
Here, $p_j = [p_{x,j}, p_{y,j}, p_{z,j}]^T$ represents the center of mass for link $j \in \{1,\ldots,n\}$.
The friction parameters $\frictionparams \in \R^{4n}$ capture joint friction characteristics:
\begin{equation}
\frictionparamsj =[F_{c_{j}}, F_{v_j}, I_{a_j}, \beta_j]^T,
\end{equation}
where these components model static friction, viscous friction, transmission inertia, and bias for each joint $j$.


This work addresses the identification of end-effector dynamics parameters, including attached payloads, while accounting for uncertainties in the complete robotic system. 
The dynamic parameters naturally partition into two components: the end-effector parameters $\inertialparamsend \in \R^{10}$ and the remaining robot parameters $\inertialparamsrobot \in \R^{14n - 10}$.
In practical applications, our objective extends beyond point estimates to establishing conservative bounds on the end-effector's dynamic parameters that provably contain the true parameters. 
This challenge is compounded by uncertainties in the base robot's parameters, leading to the following assumption:
\begin{assum}[Dynamics Parameter Bounds] \label{ass:theta-r}
The robot dynamic parameters excluding the end-effector, $\inertialparamsrobot$, lie within a known finite interval $\intparamsrobot$ containing the true parameters $\trueparamsrobot$. 
Furthermore, a nominal estimate $\nomparamsrobot \in \intparamsrobot$ of these parameters is available.
\end{assum}
\noindent This assumption aligns with practical robotics applications, as manufacturers typically provide baseline kinematic and inertial parameters for each robot link, including the unloaded end-effector. 
Friction parameters can be characterized through established methods when the robot operates without a payload, following approaches detailed in \cite{paper-constrained} and \cite{paper-iterative-sysid}. 
The interval bounds $\intparamsrobot$ can then be constructed by expanding around the nominal parameters $\nomparamsrobot$ to account for manufacturing uncertainties.

\subsubsection{Equation of Motion}
The \emph{classical equation of motion} takes the form:
\begin{multline} \label{eq:CEoM}
    H(\q, \theta) \ddq + C(\q, \dq, \theta) \dq + g(\q, \theta) \\
    + F(\dq, \ddq, \theta) = \tau(t),
\end{multline}
where $H(\q, \theta) \in \R^{n\times n}$ is the positive definite \defemph{inertia mass matrix},
$C(\q, \dq, \theta) \in \R^{n\times n}$ is the \defemph{Coriolis matrix},
$g(q, \theta) \in \R^{n}$ is the \defemph{gravitational force vector}, and
$\tau(t)\in\R^{n}$ is applied joint torques  all at time $t$.
The \emph{friction model} $F(\dq, \ddq, \theta) \in \R^{n}$, follows \cite[(2)]{paper-iterative-sysid}:
\begin{multline} \label{eq:friction}
    F(\dot{q}(t), \ddot{q}(t), \theta) = F_{c} \circ \text{sign}(\dot{q}(t)) \\
    + F_{v} \circ \dot{q}(t) + I_{a} \circ \ddot{q}(t) + \beta,
\end{multline}
where $F_{c} \in \R^n$ is the \defemph{static friction coefficients}, $F_{v} \in \R^n$ is the \defemph{viscous friction coefficients}, $I_{a} \in \R^n$ is the \defemph{transmission inertias}, $\beta \in \R^n$ is the \defemph{offset/bias terms} of all joints, and
$\circ$ denotes the Hadamard product.
In addition to this model, we assume that the robot must satisfy several limits.
\begin{assum}[Workspace and Configuration Space]
The robot's $j$\ts{th} joint has position and velocity limits given by $q_j \in [\qlim^-, \qlim^+]$ and $\dot{q}_j \in [\dqlim^-, \dqlim^+]$ for all time and for all $j \in N_q$, respectively.
During online operation, the robot has encoders that allow it to measure its joint positions and velocities.
The robot's input $\tau$ has limits given by $\tau_j(t) \in [\taulim^-, \taulim^+]$ for all time and $j \in N_q$. 
\end{assum}

The classical equations of motion \eqref{eq:CEoM} can be reformulated using a momentum-based approach, which offers practical advantages for system identification:
\begin{thm}[Momentum-Based Dynamics {\cite[(5)]{paper-momentum}}]
\label{thm:momentum-dynamics}
For a robotic system with dynamics described by \eqref{eq:CEoM}, define the system momentum for any time $t$ as:
\begin{equation} \label{eq:momentum}
p(\q, \dq, \theta) = H(\q, \theta)\dq.
\end{equation}
Then
\begin{multline} \label{eq:momentum-derivative}
    \dot{p}(\q, \dq, \theta) = C^T(\q, \dq, \theta)\dq - g(\q, \theta) \\
    - F(\dq, \ddq, \theta) + \tau(t),
\end{multline}
and for any time interval $[t_1, t_2]$ with $0 \leq t_1 < t_2$, the change in momentum satisfies:
\begin{multline} \label{eq:momentum-dynamics}
    p(q(t_2), \dot{q}(t_2), \theta) - p(q(t_1), \dot{q}(t_1), \theta) = \\
    = \int_{t_1}^{t_2} (C^T(\q, \dq, \theta)\dq - g(\q, \theta) \\
    - F(\dq, \ddq, \theta) + \tau(t)) \text{d}t,
\end{multline}
\end{thm}
\noindent This momentum-based dynamics formulation provides two key advantages: it eliminates the need for acceleration measurements, which are often noisy or unavailable in practice, and it expresses the dynamics in an integral form that filters measurement noise.


\subsubsection{Environment}
We begin by defining obstacles:
\begin{assum}[Obstacles]
\label{assum:obstacles}
The transformation between the world frame of the workspace and the base frame of the robot is known, and obstacles are expressed in the base frame of the robot.
At any time, the number of obstacles $\nObs \in \N$ in the scene is finite ($\nObs < \infty$).
Let $\obsset$ be the set of all obstacles $\{ O_1, O_2, \ldots, O_{\nObs} \}$.
Each obstacle is bounded and static with respect to time.
The manipulator has access to a zonotope overapproximation \cite{guibas2003zonotopes} of each obstacle's volume in workspace.
\end{assum}
\noindent The manipulator is \defemph{in collision} with an obstacle if $\FO_j(\q) \cap O_i \neq \emptyset$ for any $j \in N_q$ or $i \in \{1,\ldots,\nObs\}$, where $\FO_j$ describes the forward occupancy of the $j$\ts{th} link of the manipulator.
This work is concerned with planning and control around obstacles while performing system identification, not perception of obstacles.
As a result, we have assumed for convenience that the robot is able to sense all obstacles in the workspace. 

\subsection{Dynamics Regressors}
\label{sec-03-prelim-04-regressors}

To enable system identification of the dynamics parameters $\theta$, we introduce two regressor formulations that each expose the linear relationship between system dynamics and these parameters. 
These regressors provide different perspectives on the system dynamics and serve distinct purposes in controller design and system identification.
\begin{thm}[Standard Dynamics Regressor {\cite[(12)]{paper-linear-relationship}}]
\label{thm:standard-regressor}
For any time $t$, the left hand side of the classical equation of motion \eqref{eq:CEoM} can be expressed as a linear function of the dynamic parameters:
\begin{equation} \label{eq:W}
W(\q, \dq, \ddq)\theta = \tau(t).
\end{equation}
 We refer to $W(\q, \dq, \ddq)\in\R^{n\times 14n}$ as the standard dynamics regressor \cite{paper-regressor}.
\end{thm}
\noindent With this linear relationship, a least squares problem can be formulated based on \eqref{eq:W} to identify the dynamic parameters \cite[(9)]{paper-iterative-sysid}.
The momentum-based dynamics (Theorem \ref{thm:momentum-dynamics}) preserves the similar linear relationship.


\begin{thm}[Momentum-Based Regressors {\cite[(8)]{paper-momentum} and \cite[Appendix]{paper-CTq-regressor}}]
\label{thm:momentum-regressors}
The momentum dynamics admit two specialized regressors for any time $t$:
\begin{enumerate}
\item The system momentum \eqref{eq:momentum-dynamics} can be expressed as:
\begin{equation}
p(\q, \dq, \theta) = W_m(\q, \dq)\theta,
\end{equation}
where $W_m(\q, \dq) \in \R^{n\times 14n}$ is the momentum regressor.
\item The momentum derivative \eqref{eq:momentum-derivative} without torque input can be expressed as:
\begin{multline}
    C^T(\q, \dq, \theta)\dq - g(\q, \theta) -F(\dot{q}(t), \ddot{q}(t), \theta)= \\
    = W_c(\q, \dq)\theta - I_a \circ \ddq,
\end{multline}
where $W_c(\q, \dq) \in \R^{n\times 14n}$.
\end{enumerate}
Combining these regressors yields the integral form for $t_2 > t_1$:
\begin{multline} \label{eq:momentum-dynamics-integration}
    \big(W_m(q(t_2), \dot{q}(t_2)) - W_m(q(t_1), \dot{q}(t_1))\big)\theta = \\
    \left(\int_{t_1}^{t_2} W_c(\q, \dq) \text{d}t\right) \theta -  \\
    I_a \circ (\dot{q}(t_2) - \dot{q}(t_1)) + \int_{t_1}^{t_2}\tau(t)\text{d}t
\end{multline}
\end{thm}
Similarly, a least squares problem can be formulated using \eqref{eq:momentum-dynamics-integration} to identify the dynamic parameters \cite[(16)]{paper-momentum}.
However, in this instance, one can rely on just position and velocity measurements rather than acceleration measurements. 
Because this work focuses on identification of the end-effector, we separate the regressors into two parts.
In other words, we decompose $W_m$ into two components, $W_{m,r}(\q, \dq) \in \R^{n\times (14n - 10)}$ and $W_{m,e}(\q, \dq) \in \R^{n\times 10}$, where $W_{m,r}$ corresponds to the columns of $W_m$ associated with $\inertialparamsrobot$ and $W_{m,e}$ corresponds to the columns of $W_m$ associated with $\inertialparamsend$. 
We similarly decompose $W_c$ into $W_{c,r}$ and $W_{c,e}$.

\subsection{Physical Consistency Constraints and Parameterization}
\label{sec-03-prelim-05-LMI}

Using the regressors described in the previous subsection, the system identification problem can be formulated as a linear regression task. 
However, to ensure the physical consistency of the inertial parameters of each link of the robot, additional mathematical constraints must be imposed \cite{Sousa2019}. 
These constraints, also known as LMI constraints and defined in Definition \ref{defn:consistency}, require that the pseudo-inertia matrix \eqref{eq:lmi}, which encodes the inertial parameters of a robot link, be positive-definite. 
Consequently, the system identification problem becomes a constrained optimization problem \cite[(20)]{Sousa2019}, where physical consistency is explicitly enforced.

An alternative approach to enforcing positive-definiteness is to utilize the Cholesky decomposition of the pseudo-inertia matrix \cite{paper-logcholesky}. 
Specifically, this structure of $\inertialparamsj$ can be fully described by 
\begin{equation} \label{eq:theta-eta} 
\inertialparamsj = P(\eta), 
\end{equation} 
where the \emph{log-Cholesky parameters} $\eta \in \R^{10}$ are defined as \begin{equation} 
\eta = \begin{bmatrix} \alpha & d_1 & d_2 & d_3 & s_{12} & s_{13}& s_{23}& s_{14}& s_{24}& s_{34} 
\end{bmatrix}^T, 
\end{equation} 
and the mapping $P: \R^{10} \rightarrow \R^{10}$ is given by 
\begin{equation} \label{eq:P}
P(\eta) = e^{2\alpha}
    \begin{bmatrix}
    s_{14}^2 + s_{24}^2 + s_{34}^2 + 1 \\
    s_{14} e^{d_1} \\
    s_{14} s_{12} + s_{24} e^{d_2} \\
    s_{14} s_{13} + s_{24} s_{23} + s_{34} e^{d_3} \\
    s_{12}^2 + s_{13}^2 + s_{23}^2 + e^{2d_2} + e^{2d_3} \\
    s_{13}^2 + s_{23}^2 + e^{2d_1} + e^{2d_3} \\
    s_{12}^2 + e^{2d_1} + e^{2d_2} \\
    - s_{12} e^{d_1} \\
    - s_{12} s_{13} - s_{23} e^{d_2} \\
    - s_{13} e^{d_1}
    \end{bmatrix}.
\end{equation}
Further details regarding this formulation and its properties are provided in Appendix \ref{app:physical_consistency}.
\section{Robust, Online Identification For Payloads}
\label{sec-05-sysid}

This work focuses on the identification of the inertial parameters of the end-effector link, including the payload using the model and assumptions described in Section \ref{sec-03-prelim-03-dynamics}.
The rest of the section discusses how to accurately estimate $\nomparamsend$ with a conservative interval bound $\intparamsend$ that includes the true $\trueparamsend$ through a system identification process.

\subsection{System Identification Problem Formulation}
\label{sec-05-sysid-01-opt}

This subsection develops the mathematical framework for identifying end-effector parameters from measurement data. 
The approach leverages momentum-based dynamics and builds toward an optimization problem that ensures physical consistency.
\begin{defn}[Measurement Data]
For system identification, we consider $N$ sequential measurements of the robot state, collected at times $t_i \geq 0$ for $i\in{1,\ldots,N}$ that we refer to as the \emph{measurement data}:
\begin{align}
    \mathbf{q}_{1:N} &= \{q(t_i)\}_{i=1}^N \\
    \dot{\mathbf{q}}_{1:N} &= \{\dot{q}(t_i)\}_{i=1}^N \\
    \boldsymbol{\tau}_{1:N} &= \{\tau(t_i)\}_{i=1}^N.
\end{align}
\end{defn}
\noindent Using this measurement data and the momentum regressor described in Theorem \ref{thm:momentum-regressors}, one can prove the following corollary:
\begin{cor}[Momentum-Based Parameter Identification]
\label{cor:parameter_id}
Let $h \in \mathbb{N}^+$ denote the forward integration horizon. 
If one applies Forward Euler Intergration, then the momentum dynamics over interval $[t_i, t_{i+h}]$ yields a linear relationship with respect to the end-effector's dynamics parameters:
\begin{equation}
Y_{i : i + h} \trueparamsend = U_{i : i + h},
\end{equation}
where the regressor matrix $Y_{i : i + h}$ captures the end-effector dynamics:
\begin{multline} \label{eq:regressor-Yi}
Y_{i : i + h} = (W_{m, e}(q(t_{i + h}), \dot{q}(t_{i + h})) - W_{m, e}(q(t_i), \dot{q}(t_i))) - \\
- \sum_{j = i}^{i + h} W_{c, e}(q(t_j), \dot{q}(t_j))(t_{j + 1} - t_{j}),
\end{multline}
and $U_{i : i + h}$ describes the rest of the system's dynamics:
\begin{multline} \label{eq:regressor-Ui}
U_{i : i + h} =- (W_{m, r}(q(t_{i + h}), \dot{q}(t_{i + h})) - W_{m, r}(q(t_i), \dot{q}(t_i)) + \\
+ \sum_{j = i}^{i + h} W_{c, r}(\q, \dq)(t_{j + 1} - t_{j}))\theta_r - \\
- I_a \circ (\dot{q}(t_{i + h}) - \dot{q}(t_i)) + \sum_{j = i}^{i + h}\tau(t_j)(t_{j + 1} - t_{j}).
\end{multline}
\end{cor}

\noindent By applying this corollary, one can show that the system identification problem can be formulated as the solution to an optimization problem when there is no uncertainty in the robot dynamics:
\begin{thm}[Uncertainty-Free System Identification]
\label{thm:uncertainty_si}
Let $h \in \mathbb{N}^+$ denote the forward integration horizon,
suppose that measurement data was generated by the system dynamics while satisfying forward Euler Integration, and suppose that the robot's dynamic parameters excluding the end-effector are known exactly and are equal to $\nomparamsrobot$.
Let the \defemph{observation matrix} $\mathbf{Y}$ be defined as follows:
\begin{equation} \label{eq:observation-regressor}
    \mathbf{Y}(\mathbf{q}_{1:N}, \dot{\mathbf{q}}_{1:N}) \vcentcolon= \begin{bmatrix}
        Y_{1 : 1 + h} \\ 
        Y_{1 + h : 1 + 2h} \\ 
        \vdots \\ 
        Y_{1 + (N_h - 1)h : 1 + N_h h}
    \end{bmatrix} \in \R^{N_h n\times 10},
\end{equation}
and let the \defemph{observation response} $ \mathbf{U}$ be defined as follows:
\begin{equation} \label{eq:observation-response}
    \mathbf{U}(\mathbf{q}_{1:N}, \dot{\mathbf{q}}_{1:N}, \boldsymbol{\tau}_{1:N}, \trueparamsrobot) \vcentcolon= \begin{bmatrix}
        U_{1 : 1 + h} \\ 
        U_{1 + h : 1 + 2h} \\ 
        \vdots \\ 
        U_{1 + (N_h - 1)h : 1 + N_h h}
    \end{bmatrix} \in \R^{N_h n},
\end{equation}
where $N_h = \lfloor \frac{N}{h} \rfloor$ is the total number of integration steps.
Consider the following optimization problem:
\begin{equation} \label{eq:sysid-opt2}
    \min_{\eta_e \in \R^{10}} \lVert\mathbf{Y}(\mathbf{q}_{1:N}, \dot{\mathbf{q}}_{1:N})P(\eta_e) - \mathbf{U}(\mathbf{q}_{1:N}, \dot{\mathbf{q}}_{1:N}, \boldsymbol{\tau}_{1:N}, \nomparamsrobot)\rVert,
\end{equation}
where $P$ is the Log-Cholesky parameterization defined in \eqref{eq:P}. 
Let $\eta_e^*$ denote any local minimizer to this optimization problem and let $\theta_e^* = P(\eta_e^*)$, then $\mathbf{Y}(\mathbf{q}_{1:N}, \dot{\mathbf{q}}_{1:N}) \theta_e^* =  \mathbf{U}(\mathbf{q}_{1:N}, \dot{\mathbf{q}}_{1:N}, \boldsymbol{\tau}_{1:N}, \nomparamsrobot)$.
\end{thm}
\begin{proof}
This result follows by first recognizing that in the presence of no uncertainty, one can formulate the system identification problem as:
\begin{align} \label{eq:sysid-opt1}
    \min_{\trueparamsend \in \R^{10}} & && \lVert\mathbf{Y}(\mathbf{q}_{1:N}, \dot{\mathbf{q}}_{1:N})\trueparamsend - \mathbf{U}(\mathbf{q}_{1:N}, \dot{\mathbf{q}}_{1:N}, \boldsymbol{\tau}_{1:N}, \nomparamsrobot)\rVert \nonumber \\
    \text{s.t.} & && \text{LMI}(\trueparamsend) \succ \mathbb{0}_{4\times 4},
\end{align}
where have applied Definition \ref{defn:consistency} and Corollary \ref{cor:parameter_id}.
The result then follows by applying Corollary \ref{cor:parameter-mapping}. 
\end{proof}
The aforementioned Theorem makes several assumptions that make its practical application challenging.
First, as mentioned in Section \ref{sec-03-prelim-03-dynamics}, usually the robot's dynamic parameters excluding the end-effector are not known exactly.
Second, the measurement data may not be noise free.
Finally, the measurement data may not be generated by performing Forward Euler Integration. 
The next subsection describes how we deal with the first two challenges while generating a conservative estimate of the end-effector's dynamics parameters.

\subsection{Interval Bound Estimation via Perturbation Analysis}
\label{sec-05-sysid-02-bound}

To construct a conservative bound on the end-effector's dynamics parameters, we begin by making the following assumptions regarding the noise in the measurement: 
\begin{assum} \label{ass:perfect-pos-vel}
    The joint position and velocity measurements are perfectly accurate.
    The torque measurement is noisy, but its uncertainty can be bounded by an interval vector that we denote by $[\boldsymbol{\tau}_{1:N}] \in \mathbb{I}\mathbb{R}^{Nn}$.
\end{assum}
\noindent To understand why this assumption is reasonable, recall that the joint position and velocity of robotic arms are usually captured by optical encoders \cite{almurib2012review}.
The measurement error for such sensors arises due to the resolution of the coded disk inside the sensor \cite[Section 2.4]{li2019common}.
The noise in such sensors is small in modern robots \cite[Section IV]{paper-joint-encoder}.
On the other hand, the torque sensor is usually based on electrical and optical techniques \cite{paper-torque-sensor} that can introduce a non-negligible error.
As a result, our system identification techniques focus on dealing with the measurement error of the torque sensor $\boldsymbol{\tau}_{1:N}$ and the uncertainty associated with the dynamics parameters of the robot $\nomparamsrobot$ (excluding the end-effector).

The goal of this subsection is to understand how these various forms of uncertainty impact the system identification process. 
Using the aforementioned assumption, we simplify our notation to focus on this objective. 
We concatenate the variables $\boldsymbol{\tau}_{1:N}$ and $\nomparamsrobot$ that affect the estimation and denote this as a measurement vector $\meas \in \R^{Nn + 14n - 10}$.
We denote the uncertainty associated with this measurement vector by $[\meas] \in \mathbb{I}\R^{Nn + 14n - 10}$, which can be computed by applying Assumptions \ref{ass:theta-r} and \ref{ass:perfect-pos-vel}. 
Because the focus of this subsection is understanding how uncertainty in this measurement vector impacts the estimation of the end-effector's dynamics parameters, we denote the observation matrix by $\mathbf{Y}$ and the observation response matrix by $\mathbf{U}(\meas)$.

Using these definitions, we can conservatively bound the end-effector's dynamics parameter estimates that one generates by applying Theorem \ref{thm:uncertainty_si}:
\begin{thm}
\label{thm:errorbound}
Let $\eta_e^*(\meas)$ be any local minimizer to \eqref{eq:sysid-opt2}.
Let $\theta_e^*(\meas) = P(\eta_e^*(\meas))$, where $P$ is the Log-Cholesky parameterization defined in \eqref{eq:P}. 
Then the true dynamics parameters of the end-effector satisfy the following inclusion
    \begin{equation} \label{eq:interval-estimation}
        \theta_e \in \theta_e^*(\meas) + \frac{\partial \theta^*_e}{\partial \meas}([\meas])([\meas] - \meas).
    \end{equation}
\end{thm}
\noindent The proof to Theorem \ref{thm:errorbound} and the formula for $\frac{\partial \theta^*_e}{\partial \meas}(\meas)$ are provided in Appendix \ref{app:errorbound_proof}.

Algorithm \ref{alg:sysid} describes how to perform system identification using Theorem \ref{thm:errorbound} given data and an initial estimate for the dynamics parameters of the robot excluding the end-effector. 
Note that the output of the algorithm does not require an initial estimate of the dynamics parameters of the end-effector.
The algorithm outputs an estimate for the dynamics parameters of the entire robot along with a conservative interval-based bound on the dynamics parameters of the entire robot. 

\begin{algorithm}[t]
\caption{$(\nomparams, \intparams) = $ {\bf SysID}$(\mathbf{q}, \dot{\mathbf{q}}, \boldsymbol{\tau}, \nomparamsrobot, \intparamsrobot, h, [\meas])$}
\begin{algorithmic}[1]
    \STATE $\boldsymbol{m} \gets (\boldsymbol{\tau}, \nomparamsrobot)$

    \STATE $\mathbf{Y}(\mathbf{q}, \dot{\mathbf{q}}) \gets$  \eqref{eq:observation-regressor}

    \STATE $\mathbf{U}(\mathbf{q}, \dot{\mathbf{q}}, \boldsymbol{m}) \gets$ \eqref{eq:observation-response}

    \STATE $\eta_e^*(\boldsymbol{m}) \gets \eqref{eq:sysid-opt2}$ 

    \STATE $\theta_e^*(\boldsymbol{m}) \gets P(\eta^*(\boldsymbol{m}))$

    \STATE $ \intparamsend \gets \theta_e^*(\meas) + \frac{\partial \theta^*_e}{\partial \meas}([\meas])([\meas] - \meas)$ using Theorem \ref{thm:errorbound}

    \STATE $\nomparams \gets $ ($\nomparamsrobot$, $\theta_e^*(\boldsymbol{m})$) 

    \STATE $\intparams \gets $ ($\intparamsrobot$, $\intparamsend$) 

    \STATE {\bf Return:} ($\nomparams$, $\intparams$)

\end{algorithmic}
\label{alg:sysid}
\end{algorithm}

\section{Provably-safe Locally Exciting Trajectory Generation}
\label{sec-06-trajopt}

The objective of this paper is to perform system identification during online operation while ensuring safety. 
The previous section describes how to apply optimization to the collected data to conservatively identify the dynamics parameters of the end-effector. 
However, it does not describe how to collect the data in a safe manner. 
To solve this problem, this paper relies on Autonomous Robust Manipulation via Optimization with Uncertainty-aware Reachability (ARMOUR) \cite{michaux2023cant}, which is an optimization-based motion planning and control framework that can ensure the safety of synthesized trajectories even in the presence of model uncertainty. 
This section provides a brief overview of ARMOUR and how it must be modified to design trajectories that can make the identification of the true dynamics parameters of the end-effector occur more rapidly. 




\subsection{An Overview of ARMOUR \cite{michaux2023cant}}

ARMOUR plans safe trajectories in a receding horizon fashion that minimize a user-specified cost.
To accomplish this goal, ARMOUR optimizes over a space of possible desired trajectories that are chosen from a prespecified continuum of trajectories, with each determined by a \emph{trajectory parameter}, $k \in \K$.
During each planning iteration, ARMOUR selects a trajectory parameter that can be followed without collisions despite tracking error and model uncertainty while satisfying joint and input limits.
To ensure that ARMOUR is capable of planning in real time, each desired trajectory is followed while the robot constructs the next desired trajectory for the subsequent step.
Next, we summarize how ARMOUR accomplishes this goal by selecting a feedback controller and a planning time and describing what trajectory optimization problem it solves.

\subsubsection{Feedback Controller} 
ARMOUR associates a feedback control input over a compact time interval $T = [0, t_f] \subset \mathbb{R}$ with each trajectory parameter $k \in \K$.
Here $\K$ represents a parameterized space that includes a variety of behaviors of the robot.
This feedback control input is a function of the nominal inertial parameters $\nomparams$, the interval inertial parameters $\intparams$, and the state of the robot; however, it cannot be a function of the true inertial parameters of the robot because these are not known.
To simplify notation, we denote the feedback control input at time $t \in T$ under trajectory $k \in \K$ by $\tau(t;k)$. 
Applying this control input to the arm generates an associated trajectory of the arm. 
These position and velocity trajectories are functions of the true inertial parameters. 
We denote the position and velocity trajectories at time $t \in T$ under trajectory $k \in \K$ under the true inertia model parameters $\theta \in \intparams$ by $q(t;k,\theta)$ and $\dot{q}(t;k,\theta)$, respectively.

\subsubsection{Timing}
Because ARMOUR performs receding horizon planning, we assume without loss of generality that the control input and trajectory begin at time $t=0$ and end at a fixed time $t_f$. 
To ensure real-time operation, ARMOUR identifies a new trajectory parameter within a fixed planning time of $t\plan$ seconds, where $t\plan < t_f$.
ARMOUR must select a new trajectory parameter before completing its tracking of the previously identified desired trajectory.
If no new trajectory parameter is found in time, ARMOUR defaults to a braking maneuver that brings the robot to a stop at time $t = t_f$.

\subsubsection{Online Trajectory Optimization}
During each receding-horizon planning iteration, ARMOUR generates a trajectory by solving a tractable representation of the following nonlinear optimization:
\begin{align}
    \label{eq:optcost}
    &\underset{k\in \K}{\text{min}} &&\texttt{cost}(k) \\
    &&& q_j(t; k, \theta) \in [\qlim^-, \qlim^+]  &\forall t \in T, \theta \in [\theta], j \in N_q \nonumber \\
    &&& \dot{q}_j(t; k, \theta) \in [\dqlim^-, \dqlim^+]  &\forall t \in T,  \theta \in [\theta], j \in N_q \nonumber\\
    &&& \tau_j(t;k) \in [\taulim^-, \taulim^+]  &\forall t \in T,  \theta \in [\theta], j \in N_q \nonumber\\
    &&& \FO_j(q(t; k,\theta)) \bigcap \obsset = \emptyset  &\forall t \in T,  \theta \in [\theta], j \in N_q \nonumber
\end{align}
The cost function \eqref{eq:optcost} specifies a user-defined objective, such as bringing the robot close to some desired goal.
Each of the constraints guarantee the safety of any feasible trajectory parameter as we describe next.
The trajectory must be executable by the robot, which means the trajectory must not violate the robot's joint position, velocity, or input limits (i.e., the first three constraints in the optimization problem, respectively). 
 These constraints must be satisfied for each joint over the entire planning horizon despite model uncertainty; 
The robot must not collide with any obstacles in the environment (i.e., the last constraint in the optimization problem).
ARMOUR solves this optimization problem in a provably safe fashion in real-time despite model uncertainty \cite[Lemma 22]{michaux2023cant}.

We make one final observation about ARMOUR.
Recall that during receding horizon planning, ARMOUR must select a new trajectory parameter before completing its tracking of the previously identified desired trajectory.
Note that to do this, naively one may assume that one needs to know the future state of the manipulator to compute a trajectory parameter using ARMOUR. 
Because there is uncertainty in the system model, knowing the future state of the robot perfectly is untenable. 
Fortunately, ARMOUR does not require knowledge of the future state of the manipulator to ensure that the robot stays persistently safe.
In fact, to ensure that the manipulator is persistently safe using ARMOUR in a receding horizon fashion, one only needs to know the previously computed ARMOUR trajectory parameter \cite[Remark 12]{michaux2023cant}.

In the remainder of the paper, $\texttt{ARMOUR}$ represents a function that solves the optimization problem \eqref{eq:optcost}, denoted as:
\begin{equation}
k^* = \texttt{ARMOUR}(k_{p}, \obsset, \texttt{cost},t\plan,T, \theta_0, [\theta]),
\end{equation}
where $k^*$ represents the optimal trajectory parameters, and the inputs specify the previously computed trajectory parameter $k_{p}$, obstacle set $\obsset$, objective function $\texttt{cost}$, planning time $t\plan$, trajectory duration $T$, nominal dynamics parameters $\theta_0$, and dynamics parameter bounds $[\theta]$.
Note during the first planning iteration, one can just pass in the actual initial state of the robot instead of $k_p$.


\subsection{Modify Cost Function for Locally Exciting Trajectories} \label{ssec:exciting-cost}

This subsection describes how to design ARMOUR's cost function to generate exciting trajectories which can expedite the rate at which the dynamics parameters of the end-effector are identified. 
To appreciate why this is important, note that large model uncertainty may require an overly conservative representation of the forward occupancy while necessitating larger control effort to ensure safety. 
This could impede completing subsequent planning tasks.
We illustrate these challenges with poor system identification in the experiments as is described Section \ref{sec-07-exp}.

By looking at \eqref{eq:interval-estimation}, one can determine that the size of $\intparamsend$ depends on the measurements uncertainties.
Besides, the size of $\intparamsend$ also depends on the inverse of matrix $\frac{\partial^2 J}{\partial \eta_e^2}(\meas,\eta_e^*(\meas))$, which depends on the matrix $\mathbf{Y}^T\mathbf{Y}$, as shown in \eqref{eq:perturbation_eta} and \eqref{eq:D_ee}.
Unfortunately, in certain cases $\mathbf{Y}^T\mathbf{Y}$ may be close to singular, which yields a large bound for $\intparamsend$.
As a result, we would like to generate trajectories for the robot to follow that minimize the condition number of $\mathbf{Y}^T\mathbf{Y}$.
Such desired trajectories are usually referred to as \emph{exciting trajectories}.
To be more specific, we consider the 2-norm condition number of $\mathbf{Y}^T\mathbf{Y}$, which is essentially the ratio between the largest and the smallest singular values of $\mathbf{Y}^T\mathbf{Y}$ \cite[(5)]{paper-exciting-deriv}.

However, recalling the integral nature of the momentum regressor $\mathbf{Y}$ described in \eqref{eq:regressor-Yi}, it is computationally expensive to optimize the condition number of $\mathbf{Y}^T\mathbf{Y}$  because the gradient of the regressor matrix is required.
In this work, we instead minimize the condition number of the portion of the Standard Dynamics Regressor (Theorem \ref{thm:standard-regressor}) associated with the dynamics parameters of the end-effector.
Note that this is the approach taken by most prior work to generate exciting trajectories \cite{Gautierb,Venture2009,Ayusawa2017,Pressea,Bonnet2016}.

To be specific, consider the following regressor matrix:
\begin{equation} \label{eq:inverse_dynamics_regressor}
    \mathbf{W}(k) \vcentcolon= \\
    \begin{bmatrix}
        W_e(q(t_1;k), \dot{q}(t_1;k), \ddot{q}(t_1;k)) \\
        \vdots \\
        W_e(q(t_{N_s};k), \dot{q}(t_{N_s};k), \ddot{q}(t_{N_s};k))
    \end{bmatrix},
\end{equation}
where $W_e$ is the collection of columns corresponding to the dynamics parameters of the end-effector within \eqref{eq:W},
$t_i \geq 0$ for $i\in\{1,\ldots,N_s\}$ are the time instances of each sample of the desired trajectory, and $N_s$ is the number of samples.
To compute locally exciting trajectories, we set the cost function in ARMOUR equal to the condition number of  $\mathbf{W}(k)$.

To see why this works, recall that the original robot dynamics \eqref{eq:CEoM} is equivalent to the momentum version of the dynamics \eqref{eq:momentum-dynamics-integration} by taking time derivatives on both sides.
Hence, $\mathbf{W}$ can be thought of as approximating the time derivative of $\mathbf{Y}$, as they contribute linearly to \eqref{eq:CEoM} and \eqref{eq:momentum-dynamics-integration}, respectively.
Unfortunately, it is non-trivial to analytically prove that minimizing the condition numbers of these two regressor matrices is equivalent.
However, during the experiments described in Appendix \ref{app:regressor_equivalence}, we illustrate that minimizing the condition numbers of these two regressor matrices generates approximately similar behavior.

\subsection{Provably-safe Online System Identification Pipeline}

\begin{algorithm}[t]
\caption{Provably-Safe Online System Identification}
\begin{algorithmic}[1]
    \STATE {\bf Require:} $q_0\in\Q$, $\obsset$, $t\plan > 0$, $\intparams$, $\nomparams \in \intparams$, $h\in\mathbb{N}^+$, $[\meas]$ \label{alg:overall-input}


    \STATE {\bf Initialize:} $j=0$, $t_j = 0$, $\boldsymbol{\tau} = \{\}$, $\boldsymbol{q} = \{\}$, $\boldsymbol{\dot{q}} = \{\}$, and \newline \label{alg:overall-initialize-meas}
    \phantom{Initialize:} $k^*_0 = \texttt{ARMOUR}(q_0, \obsset, \texttt{cost},t\plan,T, \theta_0, [\theta])$  \label{alg:overall-firsttrajopt}
    \STATE {\bf If:} $k^*_{0} = \texttt{NaN}$, {\bf then} break \label{alg:nothinghappens}

    \WHILE{1} \label{alg:overall-while-begin}

        \STATE // Line \ref{lin:apply} executes simultaneously with Lines \ref{lin:opt} -- \ref{lin:else} //

        \STATE  {\bf Apply} $\tau(t;k^*_j)$  to robot for $t \in [t_j,t_j + t\plan]$ and append \newline \phantom {\bf Apply} input, position, and velocity of robot into $\boldsymbol{\tau},\boldsymbol{q}$, \newline \phantom {\bf Apply}
        and $\boldsymbol{\dot{q}}$, respectively \label{lin:apply} 

        \STATE  $k^*_{j+1} = \texttt{ARMOUR}(k^*_j, \obsset, \texttt{cost},t\plan,T, \theta_0, [\theta])$ \label{lin:opt}
    \STATE $(\nomparams, \intparams) =$ {\bf SysID}$(\mathbf{q}, \dot{\mathbf{q}}, \boldsymbol{\tau}, \nomparamsrobot, \intparamsrobot, h, [\meas])$ \label{alg:overall-sysid}

     \STATE  {\bf If} $k^*_{j+1} = \texttt{NaN}$, {\bf then} break
        \STATE {\bf Else}  $t_{j+1} \leftarrow t_j + t\plan$ and $j \leftarrow j + 1$ \label{lin:else}
    
    \ENDWHILE
     \STATE  {\bf Apply} $\tau(t;k^*_j)$  to robot for $t \in [t_j,t_j + t\plan]$ \label{lin:backup} 

\end{algorithmic}
\label{alg:overall}
\end{algorithm}

This subsection describes how to combine ARMOUR with the cost function described in the previous subsection with the system identification procedure described in Theorem \ref{thm:errorbound} to create the provably-safe online system identification algorithm described in Algorithm \ref{alg:overall}.

To begin, we assume that the robot has picked up a payload, which is then rigidly attached to the end-effector of the robot.
We assume that the robot starts at rest from a known initial state, $q_0$, given known obstacle configurations, $\obsset$. 
A user then specifies a planning time horizon, $t\plan$, provides an estimate of the nominal dynamics parameters of the robot, $\theta_0$ along with a conservative interval based bound on the dynamics parameters of the robot, $[\theta]$, selects a forward integration horizon $h \in \mathbb{N}^+$, and specifies a conservative error bound for the measurement vector, $[\boldsymbol{m}]$. 
Before performing any system identification, ARMOUR first tries to compute an exciting trajectory (Line \ref{alg:overall-firsttrajopt}).
If no feasible solution is found, then $\texttt{NaN}$ is returned and the robot does not move (Line \ref{alg:nothinghappens}). 

On the other hand if a feasible solution can be found, the while loop begins.
At this point, the previously computed optimal trajectory is tracked by ARMOUR's feedback controller and the behavior of the robot is saved in the measurement varaibles (Line \ref{lin:apply}).
While this is happening the next trajectory to be tracked is computed by ARMOUR (Line \ref{lin:opt}) and then system identification is performed using Algorithm \ref{alg:sysid} (Line \ref{alg:overall-sysid}). 
This is then repeated, unless ARMOUR is not able to find a feasible solution at which point ARMOUR uses the braking maneuver whose safety was verified in a previous planning iteration to bring the robot to a stop (Line \ref{lin:backup}).
In practice, the while loop is terminated if the the size of the interval estimate does not increase during subsequent iterations.

One can prove that Algorithm \ref{alg:overall} generates provably safe behavior while performing system identification. 
However, before proving that result, we make one observation. 
Recall that Theorem \ref{thm:errorbound} allows us to estimate conservatively bound the dynamics parameters of the end-effector without starting from an initial conservative estimate of these parameters. 
Unfortunately, to collect data to perform system identification using ARMOUR, we must rely on ARMOUR's feedback controller. 
This controller requires a conservative estimate of the dynamics parameters of the overall robot to ensure safe behavior. 
However, note that generating a conservative overapproximation of the range of the initial parameters of the end-effector with payload is not challenging. 
For example, the range for mass could be set from 0 to the maximum payload weight defined by the manufacturer.
The center of mass can also be bounded within the geometry of the end-effector and the payload.
The range of inertia can also be overapproximated given the range of mass and the center of mass. 

Finally by applying \cite[Lemma 22]{michaux2023cant} and Theorem \ref{thm:errorbound} one can prove the following result:
\begin{lem}[Algorithm \ref{alg:overall} Generates Safe Motion] \label{lem-safety}
Suppose $q_0$ is collision free and $[\theta]$ is an overapproximation that includes the true dynamics parameters of the robot, then the robot motion generated by Algorithm \ref{alg:overall} satisfies all the limits of the robot while staying collision-free.
\end{lem}

\section{Experiments \& Comparisons}
\label{sec-07-exp}

This section describes experiments that we conducted to evaluate the performance of the proposed algorithm.  
Our implementation has been open-sourced\footnote{\href{https://github.com/roahmlab/OnlineSafeSysID}{https://github.com/roahmlab/OnlineSafeSysID}}.

\subsection{Robotic Platform}
\label{sec-07-exp-01-robot}

We performed hardware experiments on Kinova-gen3, which is a 7 degree-of-freedom robotic arm.
Encoders and torque sensors are equipped on all joints.
The joint encoder position resolutions are 0.02 degrees for the first four joints and 0.011 degrees for the last three. 
Consequently, this work ignores the joint encoder measurement error and focuses only on the torque sensor  (Assumption \ref{ass:perfect-pos-vel}).
The frequency of control and data collection on Kinova-gen3 is not constant, but around 3.5-4 kHz.
The robot end-effector (the last link and the gripper) weighs about 1.28 kg.
The maximum payload weight of Kinova-gen3 is 2 kg for full-range continuous motion and 4 kg for mid-range continuous motion.
For safety reasons, in the hardware experiment, we restrict the payload weight to less than 4 kg.
All elements of Algorithm \ref{alg:overall} are implemented in C++ and executed on a desktop computer with an AMD Ryzen Threadripper 7980X CPU and 256 GB RAM.

\subsection{Experiment Settings}
\label{sec-exp}


\begin{figure*}[t!]
    \centering
    \includegraphics[width=1.8\columnwidth]{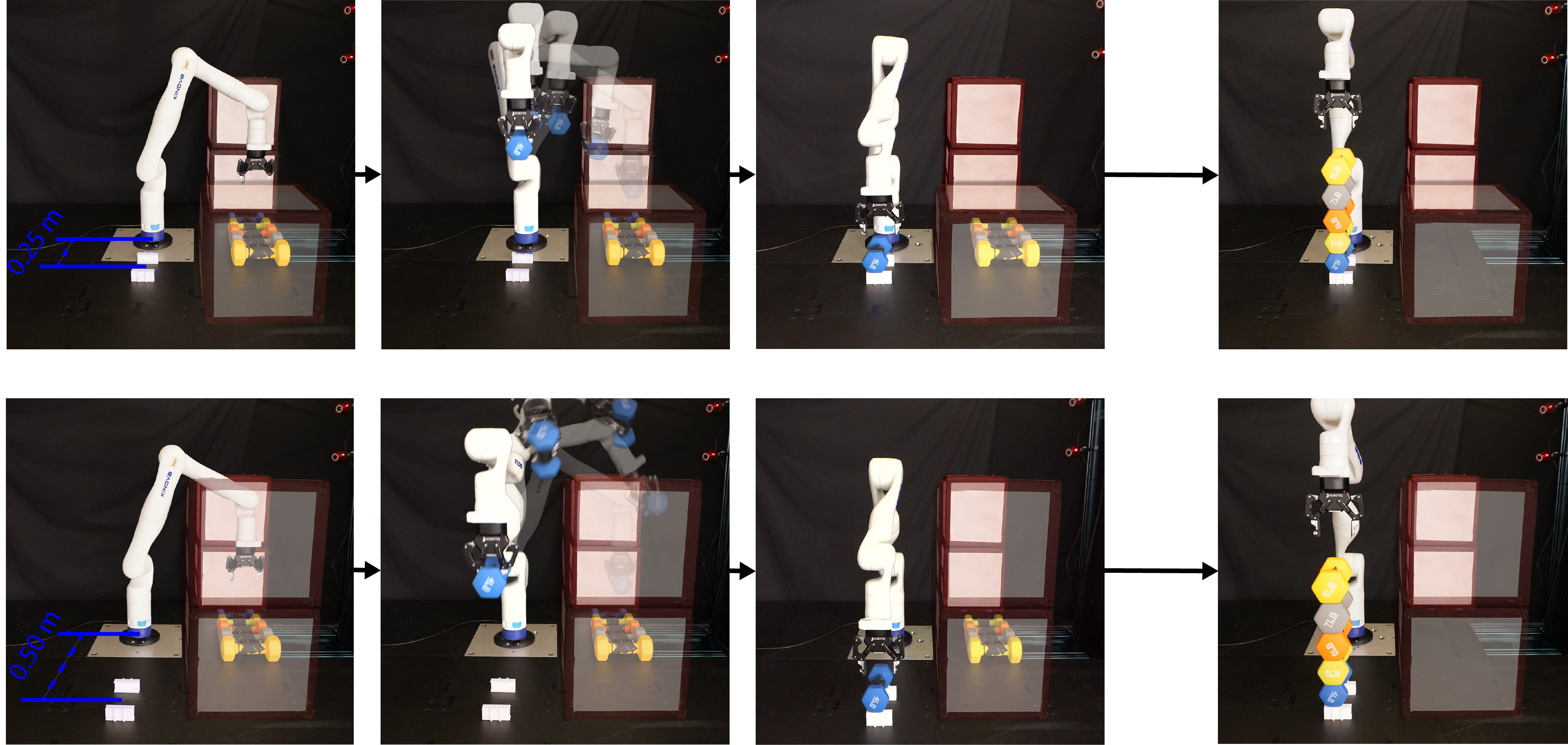}
    \caption{This figure illustrates a complex pick-and-place task used in the hardware experiment.
    Five dumbbells are placed on one side of the robot, whose inertial parameters are unknown to it (first image in both rows).
    The robot is required to move each dumbbell around the obstacles, place the lightest dumbbell on a 3D-printed platform in front of it, and stack the remaining dumbbells vertically in ascending order of weight (fourth image in both rows).
    We design two experiments with different settings:
    (a) The 3D-printed platform is positioned 0.25 m from the robot, with one low obstacle in the way, as shown in the images in the first row.
    (b) The 3D-printed platform is positioned 0.50 m from the robot, with one high obstacle in the way, as shown in the images in the second row.}
    \label{fig:task_demo}
\end{figure*}

\begin{figure}[t!]
    \centering
    \vspace*{-0.27cm}
    \includegraphics[width=\columnwidth]{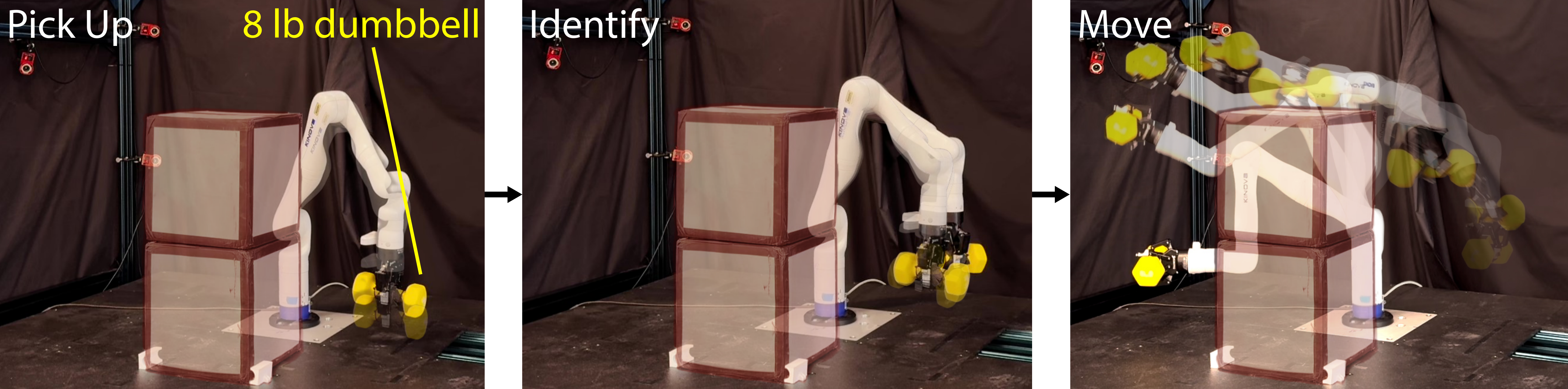}
    \caption{An illustration of the third real-world experiment.
    The robot is required to pick up and perform system identification with an 8lb dumbbell and then follow a trajectory to avoid obstacles.}
    \label{fig:expm-description}
\end{figure}

We evaluate our method across three hardware experiments:
\begin{enumerate}
    \item[(a)] Five dumbbells are placed on one side of the robot. The robot must place the lightest dumbbell on a 3D-printed platform, which is positioned at 0.25m in front of the robot, while stacking the remaining dumbbells vertically in ascending order of weight, all while avoiding obstacles in the environment, particularly with one low obstacle in the way, as illustrated at the top of Fig. \ref{fig:task_demo}.
    \item[(b)] The same stacking task is repeated with the platform positioned at 0.50m in front of the robot, particularly with one high obstacle in the way, as illustrated at the bottom of Fig. \ref{fig:task_demo}.
    \item[(c)] The robot must pick up the heaviest dumbbell (8lb) and then follow a trajectory while avoiding obstacles in a more challenging setting, as illustrated in Fig. \ref{fig:expm-description}.
\end{enumerate}
Throughout these experiments, we aim to demonstrate that our provably-safe online system identification framework is able to accomplish a variety of complex manipulation tasks with a higher success rate because it is able to provide the robust controller with a more accurate model estimate while guaranteeing safety during the system identification process.
Experiments (a) and (b) evaluate the accuracy of each method while performing manipulation of a variety of unknown payloads.
Experiment (c) evaluates the accuracy of trajectory tracking for collision avoidance.
Experiment (c) is the most challenging task because the desired trajectory is close to the obstacles to minimize the path length.
The dumbbells weigh 4, 5, 6, 7, and 8lb (about 1.81, 2.27, 2.72, 3.18, and 3.63kg, respectively).
The robot has no access to their inertial parameters in prior.

\begin{table}[ht]
    \centering
    \begin{tabular}{c|c|c}
    \hline
    \text{Inertial parameters} & \text{nominal} $\nomparamsend$ & \text{interval} $\intparamsend$ \\ 
    \hline
    $m$ (kg) & 3.2 & [1.2, 5.2] \\
    \hline
    $p_x\cdot m$ (kg$\cdot$m) & 0.0 & [-0.4, 0.4] \\
    $p_y\cdot m$ (kg$\cdot$m) & 0.0 & [-0.4, 0.4] \\
    $p_z\cdot m$ (kg$\cdot$m) & -0.5 & [-1.0, -0.1] \\
    \hline
    $XX$ (kg$\cdot$m$^2$) & 0.0 & [-0.2, 0.2] \\
    $YY$ (kg$\cdot$m$^2$) & 0.0 & [-0.2, 0.2] \\
    $ZZ$ (kg$\cdot$m$^2$) & 0.0 & [-0.2, 0.2] \\
    $XY$ (kg$\cdot$m$^2$) & 0.0 & [-0.2, 0.2] \\
    $YZ$ (kg$\cdot$m$^2$) & 0.0 & [-0.2, 0.2] \\
    $XZ$ (kg$\cdot$m$^2$) & 0.0 & [-0.2, 0.2] \\
    \hline
    \end{tabular}
    \caption{The conservative nominal parameters and interval uncertainties of the end-effector assigned to the planner and the controller at the beginning of the identification phase, which serve as initial estimates for Algorithm \ref{alg:overall} before identifying each dumbbell.
    The interval $\intparamsend$ covers the inertial parameters of the end-effector with any of the 5 dumbbells, while the nominal $\nomparamsend$ lies around the middle of $\intparamsend$.}
    \label{tab:initial_range}
\end{table}




\begin{table*}[h]
    \centering
    \begin{tabular}{c|c|c|c|c}
    \hline
    methods & controller & robot model & robot model uncertainties & exciting trajectories \\
    \hline
    \bf{ours} & \bf{ARMOUR robust \cite{michaux2023cant}} & \bf{$\theta_0$ from Algorithm. \ref{alg:sysid}} & \bf{$[\theta]$ from Algorithm. \ref{alg:sysid}} & \bf{Algorithm. \ref{alg:overall}} \\
    wrong & ARMOUR robust \cite{michaux2023cant} & assume no payload at end-effector & 0\% & N/A \\
    conservative & ARMOUR robust \cite{michaux2023cant} & nominal in TABLE~\ref{tab:initial_range} & interval in TABLE~\ref{tab:initial_range} & N/A \\
    random & ARMOUR robust \cite{michaux2023cant} & $\theta_0$ from Algorithm. \ref{alg:sysid} & $[\theta]$ from Algorithm. \ref{alg:sysid} & random \\
    adap-1 & adaptive \cite{paper-adaptive-controller} & identified by adaptive controller & N/A & N/A \\
    adap-1-excit & adaptive \cite{paper-adaptive-controller} & identified by adaptive controller & N/A & Algorithm. \ref{alg:overall} \\
    adap-2 & adaptive \cite{paper-adaptive-robust} & identified by adaptive controller & N/A & N/A \\
    adap-2-excit & adaptive \cite{paper-adaptive-robust} & identified by adaptive controller & N/A & Algorithm. \ref{alg:overall} \\
    grav-pid & gravity compensated PID \cite[(6.19)]{paper-handbook} & assume no payload at end-effector & N/A & N/A \\
    grav-pid-ours & gravity compensated PID \cite[(6.19)]{paper-handbook} & $\theta_0$ from Algorithm. \ref{alg:sysid} & N/A & Algorithm. \ref{alg:overall} \\
    grav-pid-excit & gravity compensated PID \cite[(6.19)]{paper-handbook} & $\theta_0$ from Algorithm. \ref{alg:sysid} & N/A & \cite{paper-exciting-deriv} \\
     \hline
    \end{tabular}
    \caption{
    All comparisons evaluated in the hardware experiments, categorized by the controller type, the robot model used in the controller, the robot model uncertainties used in the controller, and whether exciting trajectories were used for system identification prior to task execution.
    Methods marked "N/A" in the ``exciting trajectories'' column skip the identification phase and directly execute the task.
    }
    \label{tab:comparisons}
\end{table*}

\subsection{Comparisons}



For the hardware experiments, we evaluate a variety of comparison methods, summarized in TABLE~\ref{tab:comparisons}. 
These comparisons fall into two main categories.

The first category consists of methods that execute the task directly without an additional identification phase (``wrong'', ``conservative'', ``adap-1/2'', ``grav-pid''). 
Among these, ``wrong'' uses the same robust controller as our method, but assumes no payload is attached to the end-effector, while ``conservative'' uses a conservative estimate in TABLE \ref{tab:initial_range} that encompasses all possible dumbbells.

The second category includes methods that perform identification of the payload’s inertial parameters before executing the task (``ours'', ``random'', ``adap-1/2-excit'', ``grav-pid-ours/excit''). 
The ``random'' baseline serves as an ablation study, where identification is conducted using randomly chosen trajectories rather than the most exciting ones. 
The ``grav-pid-excit'' method utilizes a classical approach \cite{paper-exciting-deriv} based on Fourier series to generate exciting trajectories, considering only velocity and acceleration limits. 
However, these constraints may be insufficient for ensuring safety in terms of collision avoidance or torque limits when handling unknown payloads.
``adap-1/2-excit'' applies the adaptive controllers from~\cite{paper-adaptive-controller, paper-adaptive-robust} while following the same exciting trajectories as ``ours'' before executing the task, allowing the adaptive controllers additional time to estimate the end-effector inertial parameters.

\subsection{Implementation Details}

\subsubsection{System Identification Implementation}
The system identification solver (Algorithm \ref{alg:sysid}) is also implemented using Ipopt \cite{paper-ipopt}.
The analytical gradient and analytical hessian are also provided to improve computational efficiency.
The forward integration horizon $h$ is chosen to be 400, which implies a forward integration time of 100-120 ms on Kinova-gen3.
The maximum amount of uncertainties over the measurements $[\delta \meas]$ is set to 5\% maximum uncertainty for both robot dynamics parameters $\intparamsrobot$ (without end-effector) and 2.5\% for applied torque measurement $\boldsymbol{\tau}$.

\subsubsection{Exciting Trajectory Planner Implementation}
The trajectory planner \eqref{eq:optcost} is implemented using Ipopt \cite{paper-ipopt}, an open source interior point optimizer for nonlinear programming.
The duration of the exciting trajectories $t_f$ is chosen to be 3.0 s. 
The total computation time of the trajectory planner is limited to $t_p = 1.5$ s.
$N_s = 128$ samples are uniformly distributed along the trajectory to formulate the inverse dynamics regressor $\mathbf{W}$ \eqref{eq:inverse_dynamics_regressor}.
We quit the while loop in Algorithm \ref{alg:overall} after generating 4 exciting trajectories, implying a total time of $(4 - 1) \times 1.5 + 3.0 = 7.5$ s during the identification phase.

Algorithm \ref{alg:overall} requires initial estimates of the nominal parameters $\nomparamsend$ and uncertainties $\intparamsend$ of the end-effector inertial parameters.
TABLE~\ref{tab:initial_range} reports the initial nominal parameters $\nomparamsend$ and the initial interval uncertainties $\intparamsend$ at the beginning of the identification phase.
We select a range for mass that includes both the robot end-effector's mass and any dumbbell's mass. 
Assuming the robot consistently picks up the dumbbell at its center, the center of mass of the end-effector remains near the z-axis of its inertial frame. 
Thus, we select a relatively narrow range centered on 0 for $p_x\cdot m$ and $p_y\cdot m$. 
For other inertia parameters, we determine the ground truth using CAD estimation and select a range that includes the results of all dumbbells.
Note that the distribution that we use in TABLE~\ref{tab:initial_range} is much wider than that of the existing work \cite[Section IV]{paper-sensitivity}.

For the ``random" comparison, we run a different version of Algorithm \ref{alg:overall}.
Instead of optimizing for the most exciting trajectories in line \ref{lin:opt}, we randomly sample the trajectory parameter $k^*_{j+1}$ in the parameter space $\K$.

\subsubsection{Trajectory for Executing Tasks}
For all three experiments, each compared method (including our own) is required to follow a pre-defined trajectory to move the payload at a user-specified terminal location.
We compute this trajectory offline by minimizing the jerk while avoiding obstacles and satisfying joint and torque limits. 
This is done by using RAPTOR \cite{paper-RAPTOR}, an open-source trajectory optimization toolbox, with the estimated 8lb dumbbell model attached to the end-effector. 
The fast trajectory spans 2s and consists of two 1-second piecewise-continuous degree-5 Bezier curves.

\subsection{Results}
\label{sec-07-exp-04-compare}

We repeat all the experiments 5 times using the method proposed in this paper and each of the aforementioned comparison methods in TABLE~\ref{tab:comparisons}.
The results of all the experiments were consistent across all 5 trials and are summarized in TABLE~\ref{tab:task_completion_results}.
Only the method proposed in the paper (``ours") is able to successfully complete all three experiments. 
For safety reasons, we only ran ``grav-pid-excit'' in simulation because, in contrast to the trajectories generated by our approach, the trajectories from [7] violate safety constraints and torque limits, even with the lightest dumbbell.
The tracking performance and the commanded torque inputs of placing the dumbbells after the identification phase are illustrated in Figure \ref{fig:tracking_error_0.25} for Experiment (a) and Figure \ref{fig:tracking_error_0.50} for Experiment (b) in Appendix \ref{app:sys_id_results}.

\begin{table}[ht]
    \centering
    \begin{tabular}{c|c|c|c}
        \hline
        Methods & Experiment (a) & Experiment (b) & Experiment (c) \\
        \hline
        \bf{ours}      & \bf{success}   & \bf{success}  & \bf{success} \\
        wrong          & fail at 8lb    & fail at 8lb   & collide \\
        conservative   & fail at 8lb    & fail at 8lb   & collide \\
        random         & success        & fail at 8lb   & collide \\
        adap-1         & success        & fail at 8lb   & collide \\
        adap-1-excit   & success        & success       & collide \\
        adap-2         & fail at 8lb    & fail at 4lb   & collide \\
        adap-2-excit   & fail at 8lb    & fail at 4lb   & collide \\
        grav-pid       & success        & fail at 6lb   & collide \\
        grav-pid-ours  & success        & success       & collide \\
        grav-pid-excit & fail at 4lb    & fail at 4lb   & collide \\
        \hline
    \end{tabular}
    \caption{Results of our method and baseline comparisons across three hardware experiments.
    Each experiment was repeated five times and returned consistent results.}
    \label{tab:task_completion_results}
\end{table}

\begin{figure}[ht]
    \centering
    \includegraphics[width=0.9\columnwidth]{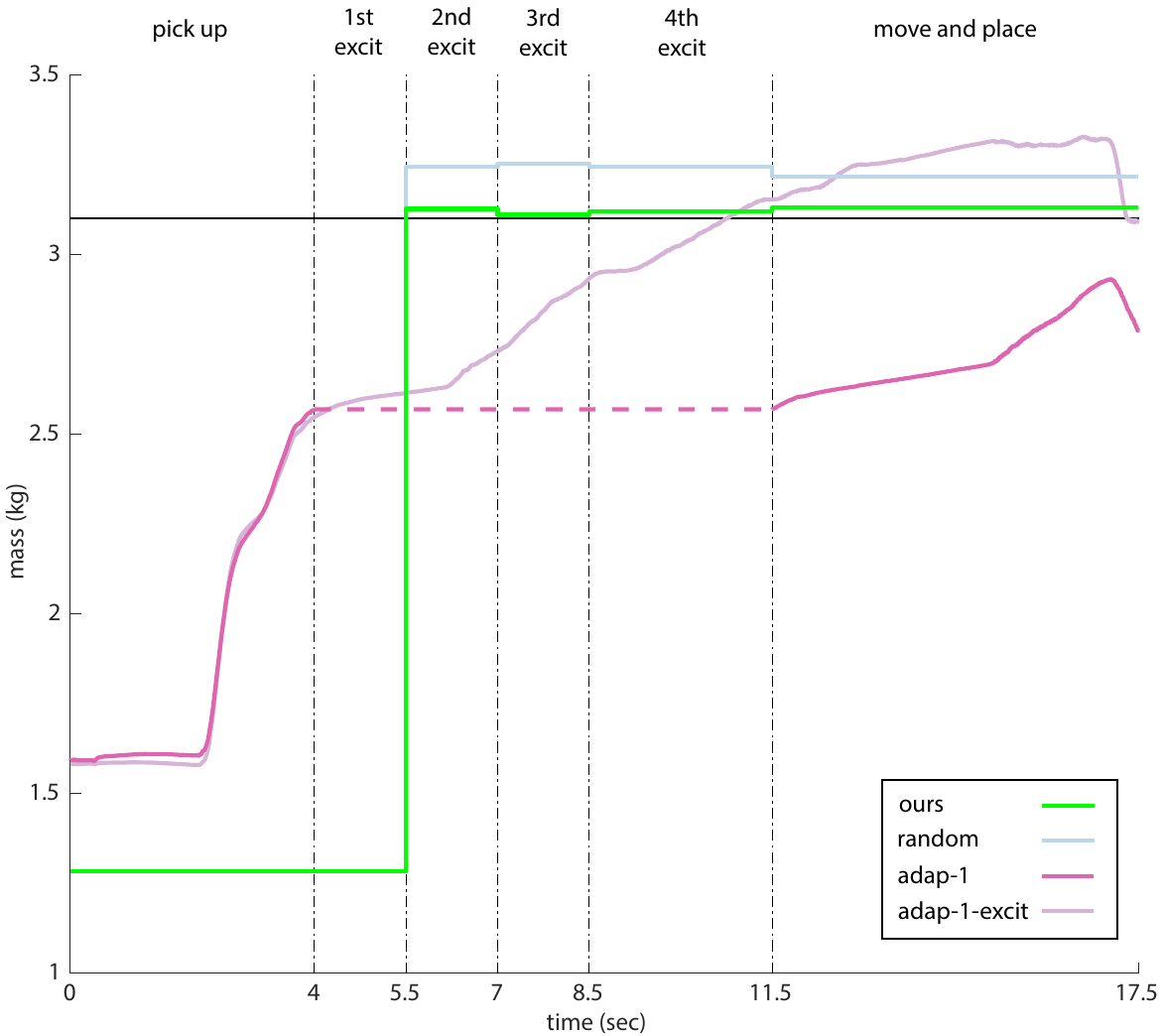}
    \caption{This figure illustrates the evolution of the estimated end-effector mass over time as the robot picks up and places the 4lb dumbbell in Experiment (b).
    For both ``ours" and ``random", we only plot the nominal estimate $\nomparamsend$.
    For both ``ours" and ``adap-1-excit", an additional identification phase (from 4 to 11.5 seconds) is included, during which the robot tracks four exciting trajectories from Algorithm \ref{alg:overall} to identify the end-effector inertial parameters.
    In contrast, ``random'' follows four randomly generated trajectories during this phase instead of exciting ones.
    For ``adap-1'', since the identification phase is skipped entirely and the robot directly executes the task (i.e., moving and stacking the dumbbells onto the platform), the identification phase (4-11.5 second) is denoted as a dotted line.
    We observe that ``ours'' achieves more accurate estimation of the end-effector parameters than random'', demonstrating the effectiveness of using exciting trajectories.
    Additionally, ``adap-1-excit'' benefits from both extended identification time and proper data excitation, resulting in improved estimation accuracy and thus, a higher success rate in the experiments compared to ``adap-1".}
    \label{fig:sysid_results_4lb_mass}
\end{figure}

Figure~\ref{fig:sysid_results_4lb_mass} shows the evolution of the estimated end-effector mass over time as the robot picks up and places the 4lb dumbbell in Experiment (b), providing further insight into the performance of methods that perform system identification of the payload’s inertial parameters before executing the task.
For both ``ours'' and ``random'', inertial parameter estimates are updated only after completing each trajectory during the identification phase (line~\ref{alg:overall-sysid} in Algorithm~\ref{alg:overall}), due to the receding horizon nature of their strategy.
In contrast, ``adap-1'' and ``adap-1-excit'' continuously update their parameter estimates from the very beginning of the experiment.
Notably, ``ours'' is more accurate while converging faster than all tested methods.
It does this before the execution of the move-and-place task, which results in a higher success rate across all experiments.
Without overburdening the readers, the results for ``adap-2'' and ``adap-2-excit'' are omitted from this figure.

\begin{figure}[h!]
    \centering
    \includegraphics[width=\columnwidth]{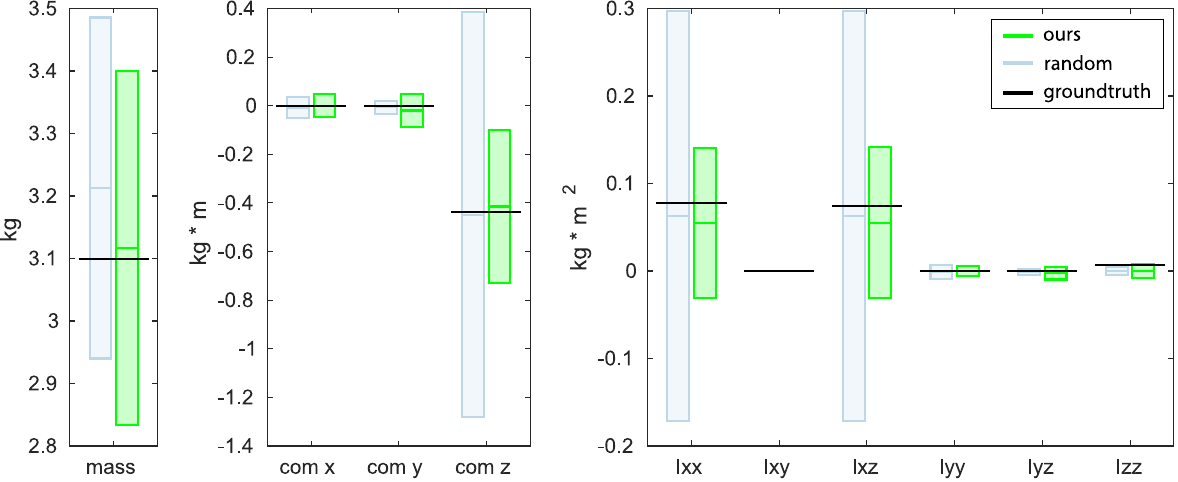}
    \caption{This figure illustrates the interval bound estimates of the 10 inertial parameters of the end-effector along with the 4lb dumbbell after the identification phase.
    Note that by following the most exciting trajectories, our method is able to achieve more accurate results and tighter interval estimates, compared to "random", which does not optimize for the most exciting trajectories.}
    \label{fig:sysid_results_4lb}
\end{figure}

Figure \ref{fig:sysid_results_4lb} reports the interval bound estimates after identifying the inertial parameters of the end-effector while manipulating the 4lb dumbbell.
Because both our method and ``random'' utilize Algorithm \ref{alg:sysid} they are guaranteed to generate conservative interval estimates for all relevant inertial parameters, which empirically validates Theorem \ref{thm:errorbound}.
In particular, given a conservative measurement noise bound, \eqref{eq:interval-estimation} can always generate overapproximated bounds that include the true inertial parameters $\trueparamsend$.
In addition, note that the interval bounds generated using the proposed method are smaller than the ones generated by selecting random trajectories for almost all inertial parameters, which illustrates that the cost function described in \eqref{eq:inverse_dynamics_regressor} is able to generate exciting trajectories.  
The identification results of 5, 6, 7, and 8lb dumbbells can be found in Appendix \ref{app:sys_id_results}.
We also report the condition numbers of the observation matrix $\mathbf{Y}$ corresponding to these results in TABLE \ref{tab:condition_numbers}.



\section{Limitations}
\label{sec-08-lim}

There exist several limitations in both the exciting trajectory planner and the robust system identification in our framework.
While most of the related work parameterize exciting trajectories as a Fourier series \cite[(50)]{paper-exciting-deriv}, the trajectory planner in our framework, which is based on ARMOUR \cite{michaux2023cant}, parameterizes the trajectories as degree-5 Bezier curves and plans for receding horizons.
Although inheriting the safety guarantees properties from ARMOUR (Lemma \ref{lem-safety}), the generated trajectories can only capture locally exciting features, generally resulting in a regressor matrix with a larger condition number than the Fourier based methods.

In addition, the approach developed in this paper has assumed that the noise in the robot manipulator system is dominated by the torque measurements.
In addition, the proposed approach assumes that the data are generated by forward Euler integration.
Though this could be extended to more accurate forms of numerical integration, these methods would still potentially just be approximations to the true system dynamics.

The Log-Cholesky parameterization $P$ \eqref{eq:theta-eta} transforms the traditional SDP formulation \eqref{eq:sysid-opt1} into an unconstrained convex problem \eqref{eq:sysid-opt2}, thereby facilitating perturbation analysis for the estimation of the interval bound of inertial parameters.
However, introducing exponentials within $P$ to enforce positive diagonal entries can also introduce numerical issues when computing the inverse of the second-order Hessian matrix in \eqref{eq:perturbation_eta} and, as a consequence, complicates the process of obtaining tighter interval bounds during online estimation.

Finally, our framework requires system identification to be performed prior to task execution, which may increase overall task completion time. 
According to our simulation experiments, trajectories that attempt to complete the task directly may lack sufficient excitation, resulting in larger interval bounds on the estimated parameters. 
These conservative estimates, in turn, lead to more conservative trajectory planning to ensure safety, ultimately prolonging task execution.

\section{Conclusion}
\label{sec-09-con}

This paper presents a provably-safe, online system identification framework for robotic arms manipulating heavy payloads with inertial uncertainties. 
By combining provably-safe trajectory optimization and robust system identification based on perturbation analysis, the framework ensures safety while refining inertial parameter estimates of the end-effector online, leading to improved planning and control performance. 
Hardware experiments demonstrate its effectiveness in handling heavy payloads under significant uncertainty, ensuring precise and safe operation. 

\bibliographystyle{plainnat}
\bibliography{references}

\appendices

\section{Physical Consistency Constraints}
\label{app:physical_consistency}

Using the regressors described in the previous subsection, one can cast the system identification problem as a linear regression problem. 
However, physical consistency of robot link parameters requires satisfying additional mathematical constraints. 
This subsection presents two approaches for ensuring these constraints: Linear Matrix Inequalities (LMI) and log-Cholesky parameterization.

\begin{defn}[Physical Consistency LMI, {\cite{Sousa2019}}]
\label{defn:consistency}
The inertial parameters of the $j\ts{th}$ link $\inertialparamsj$ are \emph{physically consistent} if they satisfy:
\begin{equation} \label{eq:lmi}
    \text{LMI}(\inertialparamsj) \vcentcolon= 
    \begin{bmatrix}
    \left(\frac{\text{tr}(I_j)}{2}\mathbb{1}_{3 \times 3} - I_j\right) \ & p_jm_j \\ p_j^Tm_j & m_j
    \end{bmatrix} \succ \mathbb{0}_{4\times 4},
\end{equation}
where $\text{LMI}(\inertialparamsj)$ is defined as \emph{pseudo-inertia matrix} for each link $j \in \{1,\ldots,n\}$,
$\text{tr}$ denotes the trace operator,
$\mathbb{1}_{3 \times 3}$ represents the 3-by-3 identify matrix,
and $I_j$ represents the \emph{inertia tensor}:
\begin{equation} \label{eq:LMI}
    I_j \vcentcolon= \begin{bmatrix}
        XX_j & XY_j & XZ_j \\
        XY_j & YY_j & YZ_j \\
        XZ_j & YZ_j & ZZ_j
    \end{bmatrix},
\end{equation}
\end{defn}
\noindent One can cast the system identification problem as trying to minimize a linear cost function while satisfying LMI constraints associated with physical consistency. 
While these LMI constraints can be incorporated into convex optimization problems {\cite[(25)]{paper-LMI}}, they require semi-definite programming (SDP) solvers and involve relaxing the LMI \eqref{eq:lmi} to positive semidefiniteness rather than strict positive definiteness.
As a result, a computed solution may not be physically consistent because the LMI may just be positive semidefinite rather than positive definite. 
An alternative approach leverages the Cholesky decomposition to ensure positive-definiteness:
\begin{thm}[Log-Cholesky Parameterization, {\cite[Corollary 7.2.9]{paper-cholesky}}]
\label{thm:log-cholesky}
Every physically consistent set of inertial parameters can be uniquely parameterized through a log-Cholesky decomposition:
\begin{equation} \label{eq:chol-decomp}
\text{LMI}(\inertialparamsj) = MM^T,
\end{equation}
where $M$ takes the form of an upper triangular matrix with positive diagonal entries:
\begin{equation}
    M \vcentcolon= e^{\alpha}\begin{bmatrix}
    e^{d_1} & s_{12}  & s_{13}  & s_{14} \\
    0       & e^{d_2} & s_{23}  & s_{24} \\
    0       & 0       & e^{d_3} & s_{34} \\
    0       & 0       &0        & 1
    \end{bmatrix}.
\end{equation}
This structure is completely characterized by the \emph{log-Cholesky parameters} $\eta \in \R^{10}$:
\begin{equation} \label{eq:eta}
\eta = \begin{bmatrix}
\alpha &  d_1 &  d_2 &  d_3 & s_{12} &  s_{13}&  s_{23}&   s_{14}&  s_{24}&   s_{34}
\end{bmatrix}^T.
\end{equation}
\end{thm}

To enforce positive definiteness and thereby enforce physical consistency of the links, one could apply the log-Cholesky decomposition \eqref{eq:theta-eta}. 
However, this would transform the convex system identification problem into a nonlinear optimization problem that may have spurious local minima. 
Fortunately, the following corollary resolves that shortcoming:
\begin{cor}[Parameter Recovery, {\cite[Section V.A]{paper-logcholesky}}]
\label{cor:parameter-mapping}
The function $P: \R^{10} \rightarrow \R^{10}$ defined in \eqref{eq:P} that transforms log-Cholesky parameters to inertial parameters is a diffeomorphism.
\end{cor}
\noindent This diffeomorphism result has significant practical implications for system identification. 
As described earlier, the log-Cholesky parameterization offers an  approach to enforce physical consistency through its structure. 
While this transformation introduces nonlinear relationships between the parameters, the diffeomorphic nature of the mapping ensures that this nonlinearity does not create spurious local minima in the optimization landscape. 
This mathematical guarantee allows us to work with the nonlinear parameterization confidently, knowing that any local minimum we find in the transformed space corresponds uniquely to a global minimum in the original parameter space.

\section{Perturbation Analysis of Convex Optimization Problems}
\label{sec-03-prelim-06-cvxpert}

The objective of this work is to understand how bounded noise in the data influences the dynamics parameter estimates.
To analyze how measurement noise affects parameter estimation in system identification, we draw upon perturbation analysis of optimization problems:

\begin{thm}[Sensitivity of Optimal Solutions, {\cite{paper-differentiability}}]
\label{thm-perturbation}
Consider a parametric optimization problem:
\begin{equation}
\min_{x\in\mathcal{X}}f(x, y),
\end{equation}
where $x$ represents the decision variable and $y$ denotes problem parameters, with $\mathcal{X}\subset\R^n$ and $\mathcal{Y}\subset\R^m$. 
If:
\begin{enumerate}
\item $f:\mathcal{X} \times \mathcal{Y} \rightarrow \R$ is continuously differentiable on both $\mathcal{X}$ and $\mathcal{Y}$
\item $f$ is geodesically convex on $\mathcal{X}$ 
and convex on $\mathcal{Y}$
\end{enumerate}
then the sensitivity of the optimal solution to parameter perturbations is given by 
\begin{equation} \label{eq:pertubation}
    \frac{\partial x}{\partial y} = -(\frac{\partial^2 f}{\partial x^2}(x, y))^{-1}\frac{\partial^2 f}{\partial x \partial y}(x, y).
\end{equation}
\end{thm}
\noindent This sensitivity analysis, discussed extensively in \cite{gould2016differentiating}, provides the foundation for bounding optimal solutions when problem parameters are subject to bounded perturbations. 
Specifically, given bounds on parameter perturbations, we can establish corresponding bounds on variations in optimal solutions.

\section{Proof of Theorem \ref{thm:errorbound}}
\label{app:errorbound_proof}

\begin{proof}
To prove \eqref{eq:interval-estimation}, first note that $\theta_e \in \theta_e^*([\meas])$. 
Hence one can apply the Mean Value Form from Interval Analysis \cite[(6.25)]{moore2009introduction} to compute the right-hand side of \eqref{eq:interval-estimation}, which describes how to overapproximate $\theta_e^*([\meas])$ by applying the Mean Value Theorem over interval arguments. 
Required by \eqref{eq:interval-estimation}, we then derive the full expression of $\frac{\partial \theta^*_e}{\partial \meas}(\meas)$.

Let $\eta_e^*(\meas)$ be any local minimizer to \eqref{eq:sysid-opt2}:
\begin{equation}
    \eta_e^*(\meas) = \argmin_{\eta_e \in \R^{10}} \lVert\mathbf{Y}P(\eta_e) - \mathbf{U}(\meas)\rVert.
\end{equation}
Note that this is also equivalent to the following form
\begin{equation}
    \eta_e^*(\meas) = \argmin_{\eta_e \in \R^{10}} J(\meas,\eta_e),
\end{equation}
where
\begin{equation}
    J(\meas,\eta_e) = \frac{1}{2} (\mathbf{Y}P(\eta_e) - \mathbf{U}(\meas))^T(\mathbf{Y}P(\eta_e) - \mathbf{U}(\meas)).
\end{equation}
The first-order optimality condition of the optimization problem above is as follows:
\begin{equation}
    \frac{\partial J}{\partial \eta_e} (\meas, \eta_e^*(\meas)) = (\mathbf{Y}P(\eta_e^*) - \mathbf{U}(\meas))^T \mathbf{Y} \frac{\partial P}{\partial \eta_e}(\eta_e^*) = 0.
\end{equation}
Treating this as an implicit equation and apply \eqref{eq:pertubation}, we can get that
\begin{equation}
\label{eq:perturbation_eta}
\frac{\partial \eta^*_e}{\partial \meas}(\meas) = -\left(\frac{\partial^2 J}{\partial \eta_e^2}(\meas,\eta_e^*(\meas))\right)^{-1} \cdot \frac{\partial^2 J}{\partial \meas \partial \eta_e} (\meas,\eta_e^*(\meas)),
\end{equation}
where 
\begin{multline} \label{eq:D_ee}
    \frac{\partial^2 J}{\partial \eta_e^2}(\meas,\eta_e^*(\meas)) = \frac{\partial P^T}{\partial \eta_e}(\eta_e^*(\meas)) \mathbf{Y}^T \mathbf{Y} \frac{\partial P}{\partial \eta_e}(\eta_e^*(\meas)) + \\
    + (\mathbf{Y}P(\eta_e^*(\meas)) - \mathbf{U}(\meas))^T \mathbf{Y} \frac{\partial^2 P}{\partial \eta_e^2}(\eta_e^*(\meas)),
\end{multline}
and
\begin{equation} \label{eq:D_em}
    \frac{\partial^2 J}{\partial \meas \partial \eta_e} (\meas,\eta_e^*(\meas)) =  - \frac{\partial \mathbf{U}^T}{\partial \meas} (\meas) \mathbf{Y} \frac{\partial P}{\partial \eta_e}(\eta_e^*(\meas)).
\end{equation}

Given $\theta_e^*(\meas) = P(\eta_e^*(\meas))$, we can then apply the chain rule:
\begin{equation}
    \frac{\partial \theta^*_e}{\partial \meas}(\meas) = \frac{\partial P}{\partial \eta_e}(\eta_e^*(\meas)) \frac{\partial \eta^*_e}{\partial \meas}(\meas),
\end{equation}
which can be further expanded using \eqref{eq:perturbation_eta}
\begin{multline}
\label{eq:perturbation_theta}
\frac{\partial \theta^*_e}{\partial \meas}(\meas) = -\frac{\partial P}{\partial \eta_e}(\eta_e^*(\meas)) \left(\frac{\partial^2 J}{\partial \eta_e^2}(\meas,\eta_e^*(\meas))\right)^{-1}  \\ \cdot \frac{\partial^2 J}{\partial \meas \partial \eta_e} (\meas,\eta_e^*(\meas)).
\end{multline}


\end{proof}

\section{Condition Number of Inverse Dynamics Regressor \eqref{eq:inverse_dynamics_regressor} and Momentum Regressor \eqref{eq:observation-regressor}}
\label{app:regressor_equivalence}

In Section~\ref{ssec:exciting-cost}, we discuss minimizing the condition number of the standard dynamics regressor $\mathbf{W}$ instead of the momentum regressor $\mathbf{Y}$ to generate exciting trajectories for system identification.
Since the original robot dynamics in \eqref{eq:CEoM} are equivalent to the momentum-based formulation in \eqref{eq:momentum-dynamics-integration} via time differentiation, $\mathbf{W}$ can be viewed as an approximation of the time derivative of $\mathbf{Y}$, as both regressors contribute linearly to \eqref{eq:CEoM} and \eqref{eq:momentum-dynamics-integration}, respectively.
Here, we numerically demonstrate that the condition numbers of these two regressors are positively correlated.

To do so, we parameterize a 10-second trajectory using a Fourier series with randomly assigned coefficients.
We sample the trajectory at a uniform time interval of 1ms, which indicates 10000 instances in total along the trajectory.
We evaluate both regressors at each time instance, and construct observation matrices from the collected data.
We then compute the condition numbers of these observation matrices for both the standard dynamics and momentum regressors across eight different values of the forward integration horizon $h$.
The positive correlation between the condition numbers of $\mathbf{W}$ and $\mathbf{Y}$ is quantified using Pearson correlation coefficients~\cite{cohen2009pearson}, as shown in TABLE~\ref{tab:correlation}.
As $h$ increases, this correlation weakens due to the accumulation of integration errors.
These results suggest that, with an appropriately chosen $h$, minimizing the condition number of the standard dynamics regressor $\mathbf{W}$ can effectively generate exciting trajectories that also reduce the condition number of the momentum regressor $\mathbf{Y}$.

\begin{table}[ht]
    \centering
    \begin{tabular}{c|c}
    \hline
    forward integration horizon $h$ & Pearson correlation coefficients \\
    \hline
    1 & 1.0000 \\
    5 & 1.0000 \\
    10 & 0.9994 \\
    20 & 0.9934 \\
    50 & 0.9540 \\
    100 & 0.6709 \\
    200 & 0.3537 \\
    500 & 0.2789 \\
    \hline
    \end{tabular}
    \caption{The Pearson correlation coefficients between the condition number of the standard dynamics regressor $\mathbf{W}$ and that of the momentum regressor $\mathbf{Y}$ are reported for different forward integration horizons $h$.
    When $h$ is small, the Pearson correlation coefficient is close to 1, indicating a strong positive correlation between the two condition numbers.
    As $h$ increases, this positive correlation gradually decreases.
    }
    \label{tab:correlation}
\end{table}



\section{Full System Identification Results}
\label{app:sys_id_results}

\begin{figure}[b!]
    \centering
    \includegraphics[width=\columnwidth]{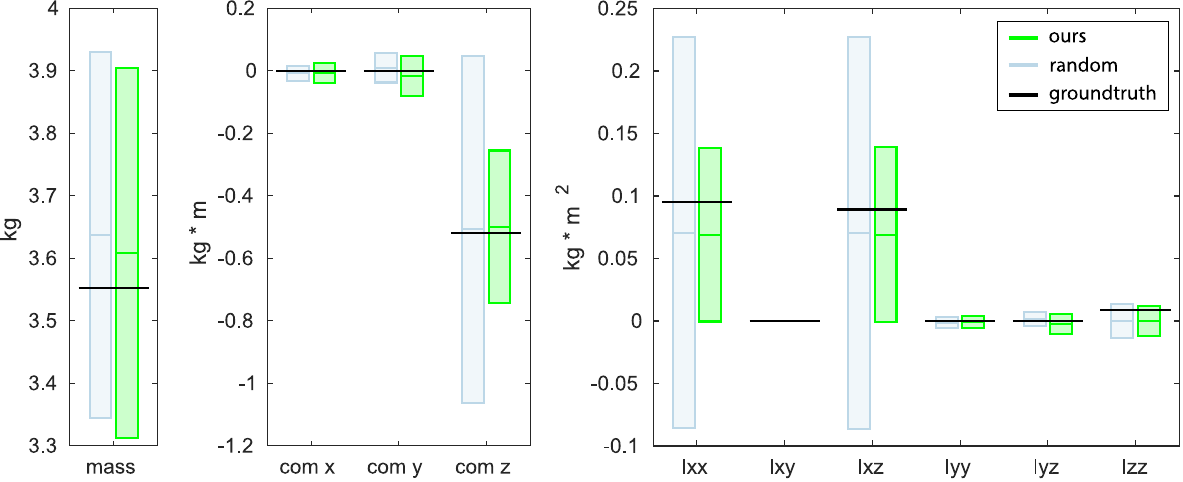}
    \caption{This figures illustrates the interval bound estimates of the 10 inertial parameters of the end-effector along with the 5lb dumbbell after the identification phase.
    Note that by following the most exciting trajectories, our method is able to achieve more accurate results and tighter interval estimates, compared to ``random", which does not optimize for excitation.}
    \label{fig:sysid_results_5lb}
\end{figure}

\begin{figure}[b!]
    \centering
    \includegraphics[width=\columnwidth]{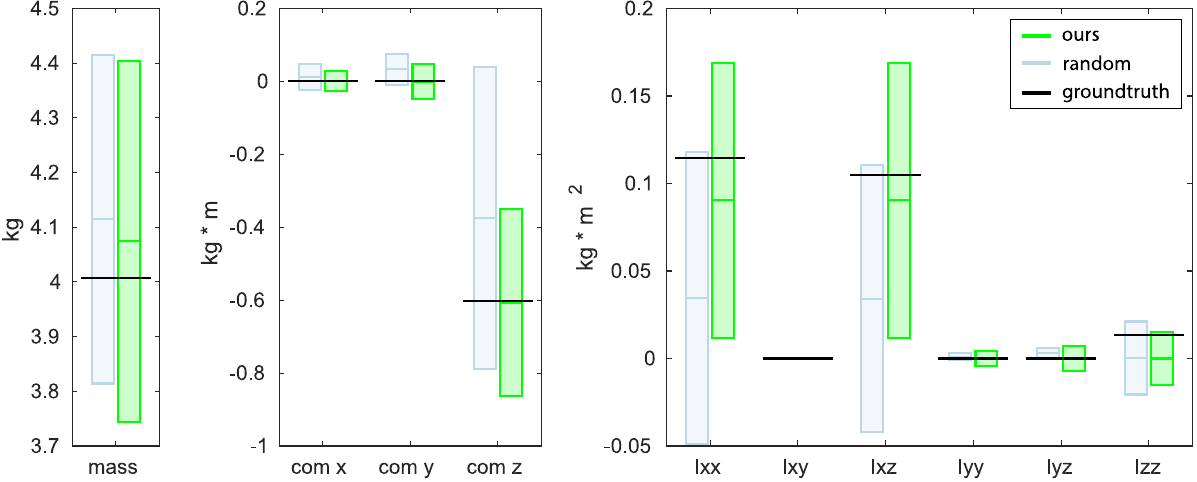}
    \caption{This figures illustrates the interval bound estimates of the 10 inertial parameters of the end-effector along with the 6lb dumbbell after the identification phase.
    Note that by following the most exciting trajectories, our method is able to achieve more accurate results and tighter interval estimates, compared to ``random", which does not optimize for excitation.}
    \label{fig:sysid_results_6lb}
\end{figure}

\begin{figure}[b!]
    \centering
    \includegraphics[width=\columnwidth]{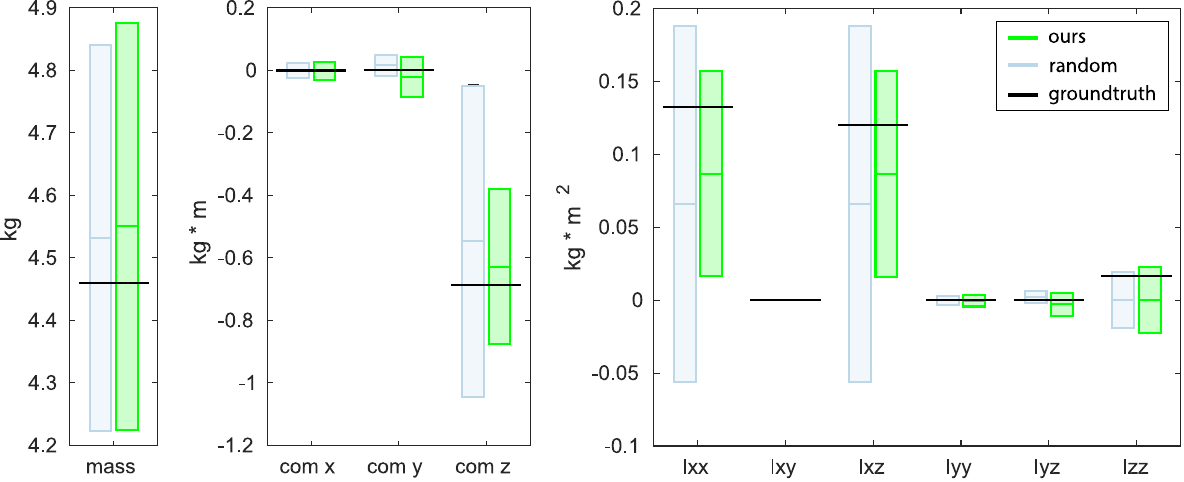}
    \caption{This figures illustrates the interval bound estimates of the 10 inertial parameters of the end-effector along with the 7lb dumbbell after the identification phase.
    Note that by following the most exciting trajectories, our method is able to achieve more accurate results and tighter interval estimates, compared to ``random", which does not optimize for excitation.}
    \label{fig:sysid_results_7lb}
\end{figure}

\begin{figure}[b!]
    \centering
    \includegraphics[width=\columnwidth]{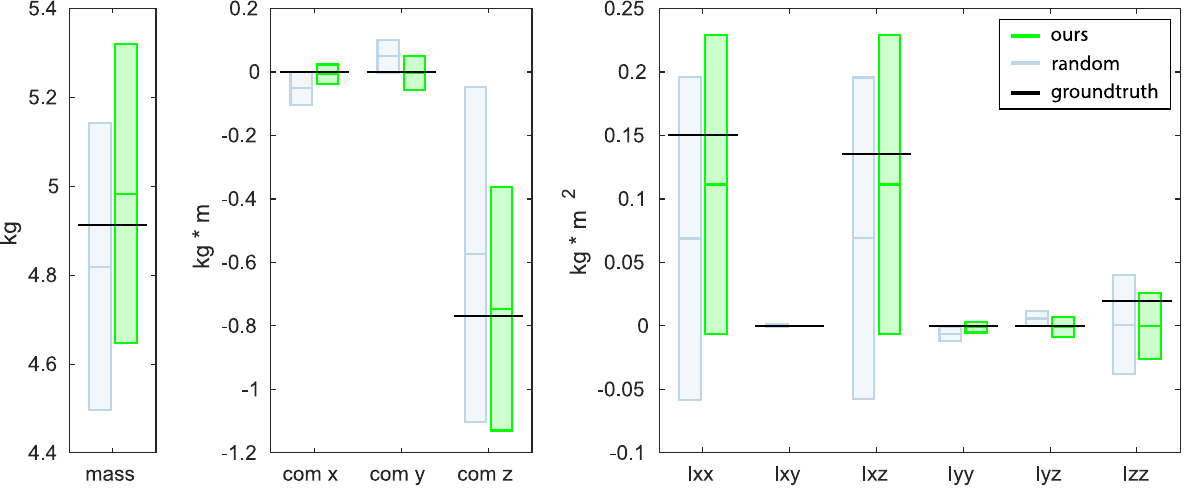}
    \caption{This figures illustrates the interval bound estimates of the 10 inertial parameters of the end-effector along with the 8lb dumbbell after the identification phase.
    Note that by following the most exciting trajectories, our method is able to achieve more accurate results and tighter interval estimates, compared to ``random", which does not optimize for excitation.}
    \label{fig:sysid_results_8lb}
\end{figure}

This section reports the full system identification results of the inertial parameters of the end-effector along with the 5, 6, 7, and 8lb dumbbells.
The final system identification results can be found in 
Figures \ref{fig:sysid_results_5lb}, \ref{fig:sysid_results_6lb}, \ref{fig:sysid_results_7lb}, and \ref{fig:sysid_results_8lb}.
We also report the condition numbers of the observation matrix $\mathbf{Y}$ corresponding to these results in TABLE \ref{tab:condition_numbers}.
Because both our method and ``Random'' utilize Algorithm \ref{alg:sysid}, they are guaranteed to generate conservative interval estimates for all relevant inertial parameters, which empirically validates Theorem \ref{thm:errorbound}.
In addition note that the interval bounds generated using the proposed method are smaller than the ones generated by selecting random trajectories for almost all inertial parameters, which illustrates that cost function described in \eqref{eq:inverse_dynamics_regressor} is able to generate exciting trajectories.

\begin{table}[ht]
    \centering
    \begin{tabular}{c|c|c}
    \hline
     dumbbell & ours & random \\
     \hline
     4lb & \bf{274.030} & 479.573 \\
     5lb & \bf{282.840} & 418.078 \\
     6lb & \bf{250.704} & 373.448 \\
     7lb & \bf{258.517} & 312.352 \\
     8lb & \bf{233.539} & 413.381 \\
     \hline
    \end{tabular}
    \caption{This table reports the 2-norm condition number of $\mathbf{Y}$ based on data collected during the entire 7.5-second identification phase. 
    By optimizing for the most exciting trajectories, our method produces an observation matrix $\mathbf{Y}$ with a significantly smaller condition number compared to the "random" baseline, which does not optimize for excitation. 
    As a result, our method achieves more accurate parameter estimation and generates tighter interval bounds, as demonstrated in the figures on the right column.
    }
    \label{tab:condition_numbers}
\end{table}

\section{Full Tracking Error \& Commanded Torque Plots}
\label{app:tracking_error_results}

\begin{figure*}[t!]
    \centering
     \includegraphics[width=2.0\columnwidth]{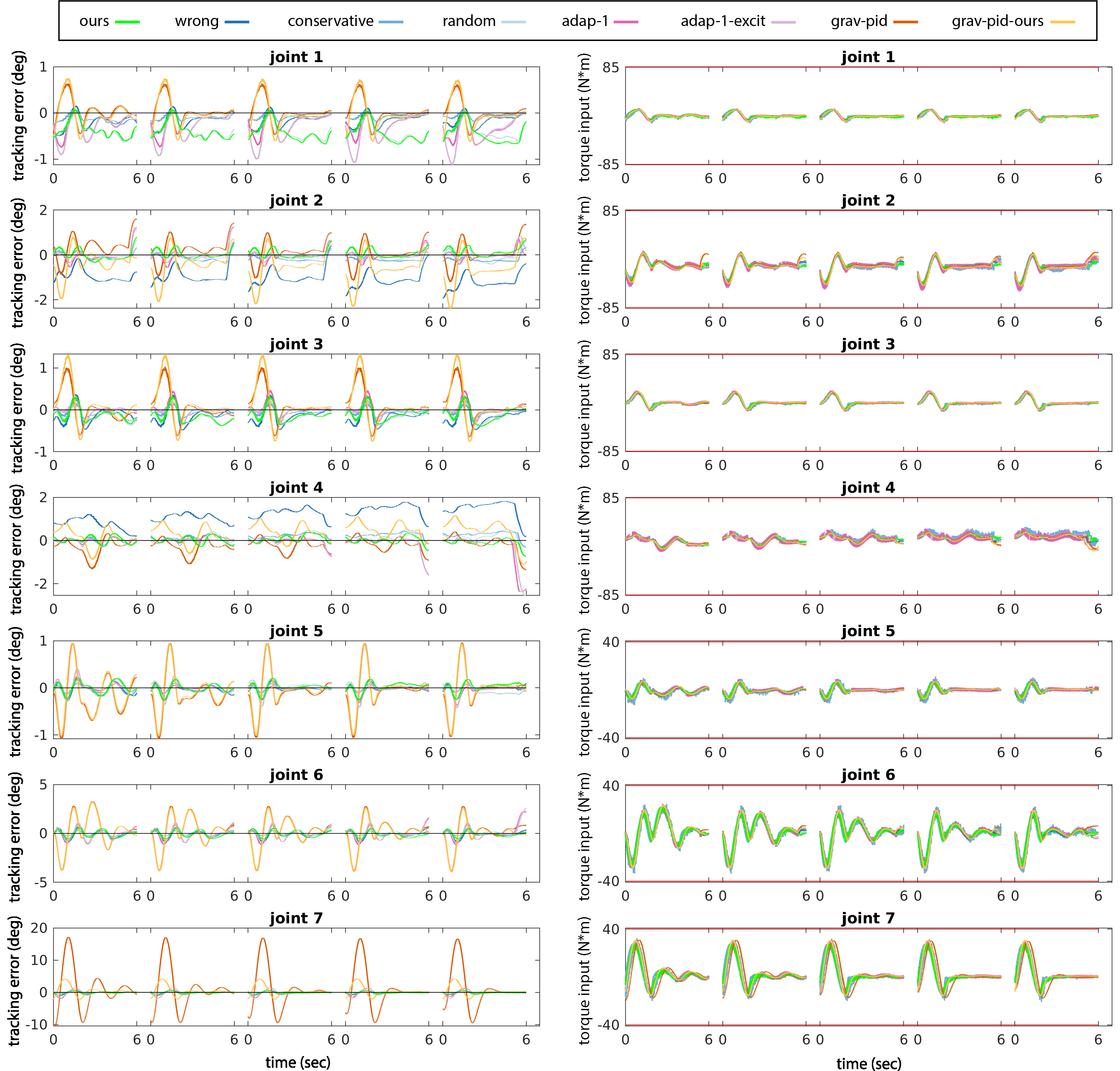}
    \caption{This figure illustrates the tracking error of our method and all the comparisons on the left and the commanded torque on the right, while moving and stacking each of the five dumbbells on the 3D-printed platform located 0.25 m in front of the robot in Experiment (a).}
    \label{fig:tracking_error_0.25}
\end{figure*}

\begin{figure*}[t!]
    \centering
    \includegraphics[width=2.0\columnwidth]{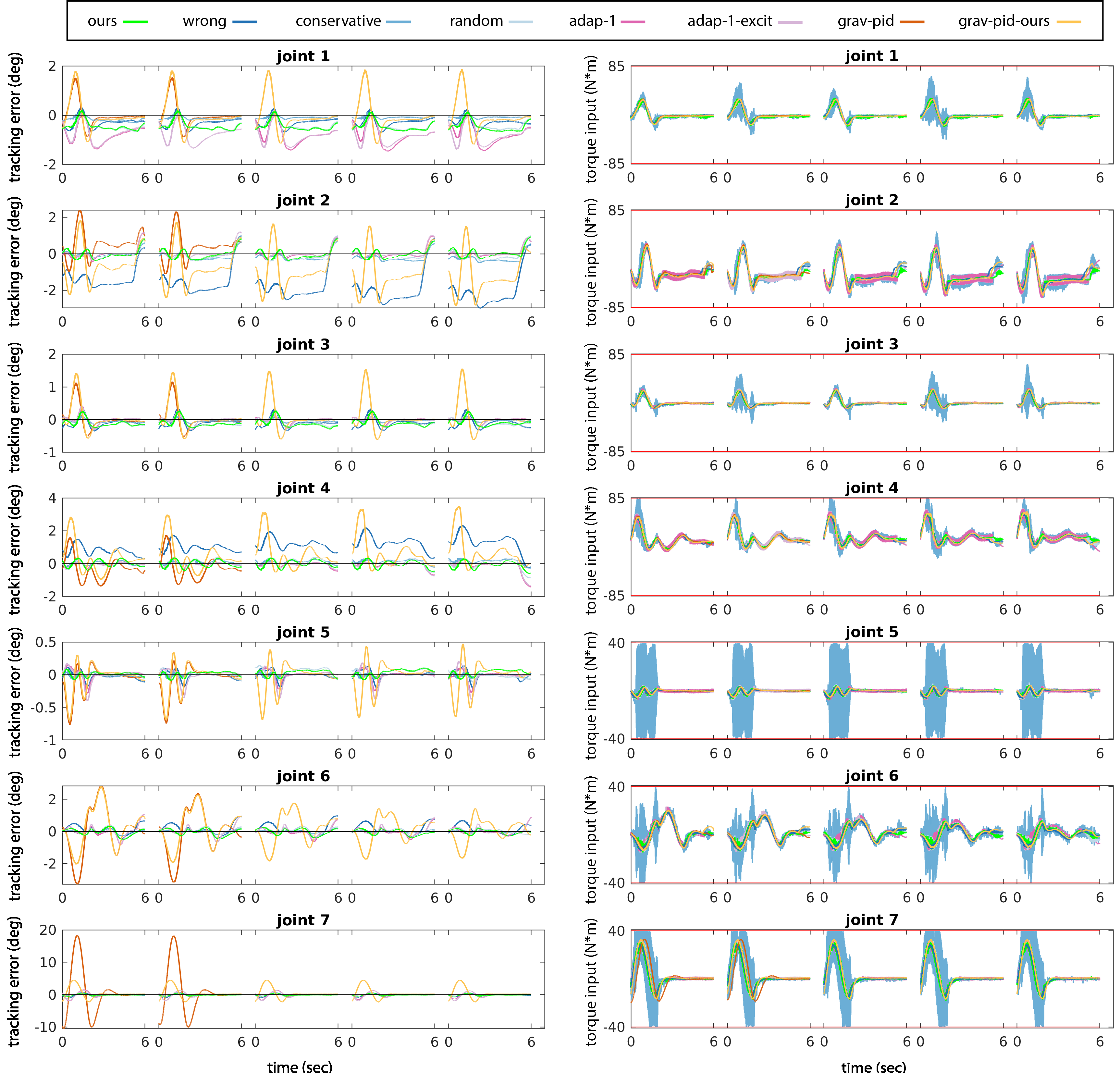}
    \caption{This figure illustrates the tracking error of our method and all the comparisons on the left and the commanded torque on the right, while moving and stacking each of the five dumbbells on the 3D-printed platform located 0.50 m in front of the robot in Experiment (b).}
    \label{fig:tracking_error_0.50}
\end{figure*}

This section reports the tracking error and the commanded torque when the robot moves each of the five dumbbells around the obstacles by following a precomputed trajectory and stacks all dumbbells vertically on a 3D-printed platform during the second phase of the second experiment.
Figure \ref{fig:tracking_error_0.25} and \ref{fig:tracking_error_0.50} include the results of our method and all the comparisons listed in TABLE \ref{tab:comparisons}.

\end{document}